\documentclass{article}

\usepackage{arxiv}

\usepackage[utf8]{inputenc} 
\usepackage[T1]{fontenc}    
\usepackage{hyperref}       
\usepackage{url}            
\usepackage{booktabs}       
\usepackage{amsfonts}       
\usepackage{nicefrac}       
\usepackage{microtype}      
\usepackage{amsmath}
\usepackage{amssymb}
\usepackage{cleveref}       
\usepackage{xcolor}
\usepackage{graphicx}
\usepackage[numbers]{natbib}
\usepackage{doi}
\usepackage{diagbox}
\usepackage{rotating}
\usepackage{adjustbox}
\usepackage{algorithm}
\usepackage{algpseudocode}
\usepackage[normalem]{ulem}
\usepackage{subcaption}
\newtheorem{remark}{Remark}

\title{ShapG: new feature importance method based on the Shapley value}

\author{ 
  \href{https://orcid.org/0000-0002-1166-7578}{\includegraphics[scale=0.06]{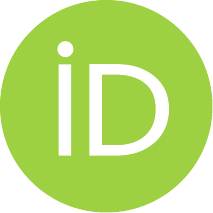}\hspace{1mm}Chi Zhao} \\ 
	Saint Petersburg State University,\\
	7/9 Universitetskaya nab.,\\
	Saint Petersburg, 199034, Russia\\
	\texttt{st081292@student.spbu.ru} \\
  \And
	\href{https://orcid.org/0000-0001-7584-7685}{\includegraphics[scale=0.06]{orcid.pdf}\hspace{1mm}Jing Liu} \\
	Saint Petersburg State University,\\
	7/9 Universitetskaya nab.,\\
	Saint Petersburg, 199034, Russia\\
	\texttt{st082130@student.spbu.ru} \\
		\And
	\href{https://orcid.org/0000-0003-3976-7180}{\includegraphics[scale=0.06]{orcid.pdf}\hspace{1mm}Elena Parilina} \\
	Saint Petersburg State University,\\
	7/9 Universitetskaya nab.,\\
	Saint Petersburg, 199034, Russia\\
	School of Mathematics and Statistics,\\
	Qingdao University,\\
	Qingdao, 266071, PR China\\
	\texttt{e.parilina@spbu.ru} \\
}

\renewcommand{\shorttitle}{ShapG: new feature importance method based on the Shapley value}

\hypersetup{
pdftitle={ShapG: new feature importance method based on the Shapley value},
pdfsubject={q-bio.NC, q-bio.QM},
pdfauthor={Chi Zhao, Jing Liu, Elena Parilina},
pdfkeywords={XAI method, ShapG, Shapley Values, Complex AI model, Feature Importance},
}

\begin{document}
\maketitle

\begin{abstract}
With wide application of Artificial Intelligence (AI), it has become particularly important to make decisions of AI systems explainable and transparent. In this paper, we proposed a new Explainable Artificial Intelligence (XAI) method called ShapG (Explanations based on Shapley value for Graphs) for measuring feature importance. ShapG is a model-agnostic global explanation method. At the first stage, it defines an undirected graph based on the dataset, where nodes represent features and edges are added based on calculation of correlation coefficients between features. At the second stage, it calculates an approximated Shapley value by sampling the data taking into account this graph structure. The sampling approach of ShapG allows to calculate the importance of features efficiently, i.e. to reduce  computational complexity. Comparison of ShapG with other existing XAI methods shows that it provides more accurate explanations for two examined datasets. We also compared other XAI methods developed based on cooperative game theory with ShapG in running time, and the results show that ShapG exhibits obvious advantages in its running time, which further proves efficiency of ShapG. In addition, extensive experiments demonstrate a wide range of applicability of the ShapG method for explaining complex models. We find ShapG an important tool in improving explainability and transparency of AI systems and believe it can be widely used in various fields.
\end{abstract}


\section{Introduction}\label{sec1}
With a widespread application of artificial intelligence in various industries such as healthcare \cite{nazar2021systematic}, finance \cite{weber2023applications}, and autonomous driving \cite{atakishiyev2021explainable}, demand for Explainable Artificial Intelligence (XAI) has become increasingly important \cite{ali2023explainable}. Machine Learning or Artificial Intelligence systems are often seen as black boxes, with opaque decision-making processes that are difficult for humans to understand \cite{von2021transparency}. This lack of transparency has raised concerns regarding trust in AI systems. Therefore, the introduction of explainable AI has become an essential choice. XAI can provide explanations for black box models and decision processes \cite{ehsan2021expanding}.

The introduction of XAI enhances users' trust in AI systems and helps them better understand and analyze the operation of models, enabling better application in real-world scenarios \cite{confalonieri2021using}. XAI can reveal the importance of different features by performing feature importance analysis or explaining decision pathways \cite{antoniadi2021current}. This allows users to understand the impact of each input feature on the model's predictions, enabling them to make informed decisions, select relevant features, or provide appropriate recommendations. Arrieta et al. reviewed concepts related to explainable AI and analyzed the types of explanations provided by XAI, primarily categorized into two types: global  and local explanations \cite{arrieta2020explainable}.

Local explanations methods in XAI focus on providing explanations for individual samples or predictions \cite{plumb2018model}, while global explanations methods provide explanations for the whole model \cite{ibrahim2019global}. These methods are highly beneficial in helping users understand the decision-making process and feature importance of a model, enhancing trust in the model, and providing explanation-based decision support. In local explanations, LIME (Local Interpretable Model-Agnostic Explanations) \cite{ribeiro2016should} and SHAP (Shapley Additive Explanations) \cite{lundberg2017unified} are two commonly used methods. In global explanations, Feature Importance and SHAP can also be used to provide explanations  \cite{lundberg2020local}.

By calculating importance of  features in ML models, we can understand the impact of different features on the model prediction results. This analysis helps us to identify  features that are crucial for prediction results, so that we can understand a model deeper and make efficient decisions. 

Explainable AI methods such as LIME (Local Interpretable Model-Agnostic Explanations) SHAP (Shapley Additive Explanations) are used on six medical digital datasets to evaluate the results confirming efficiency of integrating global and local explainability techniques and highlighting the superior performance of the global SHAP explainer \cite{hakkoum2024global}. Joo et al. analyze the effect of substances on heat deflection temperature (HDT) using SHAP to identify potential substance choices to achieve the target HDT, and the results of this study provide valuable guidance for modeling potential substance choices and adjusting substance ratios using SHAP method \cite{joo2023machine}. Lai et al. utilized SHAP methodology to assess the relative importance of input features in performance design of Concrete Filled Steel Tubes (CFSTs) under lateral impact loads and establish the relationship between input features and target outputs. In addition, they conducted a comprehensive parametric study to explore the effect of each input feature on the performance design \cite{lai2024probabilistic}. Kashifi et al. use SHAP to enhance the explainability of Gated Recurrent Convolutional Network (GRCN) models in predicting spatiotemporal crash risk, and also revealed factors related to crash risk, which can be used to develop efficient strategies for interventions to reduce crash risk and improve overall traffic safety \cite{kashifi2024robust}. Using XAI technology in non-transparent AI models, users can gain a deeper understanding of the model's predictions and make decisions based on explained results, leading to better application of AI technology in various fields.

In this paper, we proposed a new method of global explanation called ShapG (Explanations based on Shapley value for Graphs) for calculating the importance of features in machine learning models. ShapG can explain any type of model, so it is a model-agnostic XAI method. This method first constructs an undirected graph where nodes represent features and sampling is performed based on this graph.\footnote{The corresponding sampling approach to reduce the number of calculations to find Shapley value for a graph is first introduced in \cite{zhao2024investigationcentralitymeasuresopinion}.} Unlike traditional methods for calculating Shapley values, we only consider coalitions between each node and its neighbors, rather than considering all possible coalitions. Doing this, we improve the speed of calculating Shapley value, and obtain the global explanation of machine learning models. The main contributions and results of this paper are as follows:
\begin{itemize}
    \item We propose a new explainable artificial intelligence (XAI) method based on the Shapley value calculated on graphs. By applying our XAI method to two different datasets and comparing explanation results with other existing XAI methods, we have demonstrated efficiency of our method in explaining models.
    
    \item ShapG method is suitable for explaining any type of model, including complex neural network models or hybrid models. 

    \item With evaluation results of XAI, we compare and analyze that our method can give better results than  existing XAI methods in explanations and can calculate  feature importance more accurately. Meanwhile, compared with SamplingSHAP and KernelSHAP methods developed based on cooperative game theory, our method outperforms them in running time.
\end{itemize}

The structure of this paper is as follows: Section \ref{sec2} briefly describes existing XAI methods. Section \ref{sec3} describes in detail our proposed new XAI method using the  Shapley value as its basis. Section \ref{sec4} contains description of datasets, data processing based on the ShapG method, AI models and evaluation of XAI methods. Section \ref{sec5} shows and analyzes experimental results and explanation results of the complex AI models. Section \ref{sec6} briefly concludes.

\section{Explainable Artificial Intelligence (XAI) methods}\label{sec2}

In a general formulation, we assume that there is a sample of observations of features $X_1,\ldots,X_M$ which are used to construct a model $f(X_1,\ldots,X_M)$ to explain/predict/classify a target variable $Y$. Due to nontransparency of black-box models in data analysis, the functional form of $f$ is not known and there is a crucial requirement to estimate features importance in AI models applied for tabular data. There are three most popular explainable AI methods: feature importance, LIME (Local Interpretable Model-agnostic Explanations), and SHAP (SHapley Additive exPlanations). We do not describe them here but provide corresponding references where these methods are introduced or well described. 

We use following methods in comparison with our novel XAI method described in Section \ref{ShapG}:
\begin{itemize}
	\item \textbf{Feature Importance:} Feature importance is a built-in method applied to tree models such as decision trees, random forests, gradient boosting trees, etc. When constructing a tree model, the algorithm automatically calculates the contribution of each feature to the model's prediction measuring the impact of each feature on the predicted outcome and evaluating its importance \cite{louppe2014understanding}.
	
	\item \textbf{Permutation Feature Importance:} Permutation feature importance, a method for evaluating the importance of features, is proposed in \cite{altmann2010permutation}. If a feature is important, then the model performance will be greatly reduced when randomly shuffled, while if a feature is unimportant, then it will have very little impact on the model performance when it is randomly shuffled.
	
	\item \textbf{LIME:} It is a model-agnostic explanation method proposed in \cite{ribeiro2016should} and used to explain importance of variables in predictions of  machine learning models  \cite{mishra2017local}. The method is called agnostic if it is not specifically related to one particular machine learning method but can be applied to most of machine learning models.
	
	\item \textbf{SHAP (SHapley Additive exPlanations):} It is an explainable method developed based on cooperative game theory and proposed in \cite{lundberg2017unified}. SHAP provides a model-agnostic explanation mechanism that can theoretically be applied to any machine learning model. 
	
	\item \textbf{KernelSHAP:} The method is based on LIME and the Shapley value. The method is using the following steps to simplify calculations: (i) generation of a random number of samples of features, (ii) defining the sample data for each subset of features in a special way, (iii) calculation of weights for each subset of features, and (iv) solution of a specially defined weighted least square optimization problem to find the vector of features importance, that is, an approximated  Shapley value.
	
	\item \textbf{Sampling SHAP:} It computes Shapley values under the assumption of feature independence and it is an extension of the algorithm proposed in \cite{strumbelj2010efficient}. The calculations are based on a well-known alternative formulation  of the Shapley value \cite{castro2009polynomial}.
\end{itemize}

\section{ShapG: a novel XAI method}\label{sec3}

\subsection{The Shapley value}

The cooperative game is defined by $(\mathcal{M},v)$, where $\mathcal{M}=\{1,\ldots,M\}$ is the set of players\footnote{In the problem of measuring features importance, a feature is considered as a player in the game, so we use the same notation $M$ for the number of players and number  of features.} and $v: 2^{\mathcal{M}}\rightarrow \mathbb{R}$ is a characteristic function defining a ``strength'' of any coalition of players that is a subset of $\mathcal{M}$, i.e., for any coalition or collection of players $\mathcal{S}\subset \mathcal{M}$, $S=|\mathcal{S}|$. The value $v(\mathcal{S})$ represents the payoff or power of coalition $\mathcal{S}$. One of the main problems which the theory of cooperative games is solving is to find a ``fair'' allocation of the total payoff of the grand coalition $v(\mathcal{M})$ among its members. One imputation was proposed by Shapley \cite{shapley1951notes} to allocate $v(\mathcal{M})$, and it is a vector $\boldsymbol{\phi}=(\phi_1,\ldots,\phi_M)$, where $\phi_i$ is a payoff (part of $v(\mathcal{M})$) to player $i\in \mathcal{M}$ defined by
\begin{equation}\label{shap value}
\phi_i=\sum_{\mathcal{S}:\mathcal{S}\subset \mathcal{M}\setminus \{i\}} \frac{(M-S-1)!S!}{M!}\left(v(\mathcal{S}\cup \{i\})-v(\mathcal{S})\right),
\end{equation}
where $\left(v(\mathcal{S}\cup \{i\})-v(\mathcal{S})\right)$ is a marginal contribution of player $i$ if he joins coalition $\mathcal{S}$. 

The vector with the components defined by \eqref{shap value} is called the Shapley value and it is a unique vector satisfying four axioms (efficiency, symmetry, null player, and additivity). The efficiency axiom means that the sum of the components of the Shapley value is equal to the payoff of grand coalition $\mathcal{M}$, i.e. $\sum_{i\in \mathcal{M}} \phi_i=v(\mathcal{M})$.

We also provide a probabilistic interpretation (see \cite{Naumova}) of the Shapley value to better understanding why this vector can be applied to measure the feature importance in complex machine learning models. Consider the $i$-th component of the Shapley value defined by \eqref{shap value}. Player $i$ gets the payoff $\left(v(\mathcal{S}\cup \{i\})-v(\mathcal{S})\right)$ when he joins to the randomly formed coalition $\mathcal{S}:\mathcal{S}\subset \mathcal{M}\setminus \{i\}$. The probability that coalition $\mathcal{S}$ containing $S$ players is formed is equal to $\frac{1}{M}\binom{M-1}{S}$. It is assumed that all coalition $\mathcal{S}$'s sizes from $0$ to $M-1$ are equally probable and for a given coalition size $S$, the subsets of $S$ players are also equally probable. Then the value $\phi_i$ given by \eqref{shap value} is player $i$'s expected payoff in such a probabilistic scheme.

To make a connection between the Shapley value and the vector of feature importance, we can associate the set of players with the set of features and the  characteristic function with the some quantitative characteristic of the prediction made by a machine learning model using subset of features. Then, the difference $\left(v(\mathcal{S}\cup \{i\})-v(\mathcal{S})\right)$ can be interpreted as a contribution or payoff in the prediction quality if we add feature $i$ to the subset of features $\mathcal{S}$. The expected value of such payoff is associated with   feature $i$'s importance in a testing prediction ML model.

The main idea described in this section is borrowed from the theory of cooperative games and implemented in the SHAP method, but due to the complexity of the Shapley value calculation by formula \eqref{shap value} because the number of features and complexity of a prediction model $f$, the algorithms such as KernelSHAP and SamplingSHAP are proposed to approximate the Shapley value by reduction of the number of calculations  \cite{lundberg2017unified,vstrumbelj2014explaining}.

\subsection{ShapG (explanations based on the Shapley value for graphs)}\label{ShapG}

We describe a new XAI method called \textit{ShapG} to calculate the feature importance in machine learning models based on the Shapley value defined on an undirected weighted graph constructed in a special way.

\subsubsection{The Shapley value for undirected weighted graphs}

The calculation of the Shapley value for an undirected weighted graph can be divided into following steps:
\begin{enumerate}
\item We define the undirected weighted graph $G=(\mathcal{M},\mathcal{E})$, where $\mathcal{M}$ is the set of nodes which are associated with features from set $\mathcal{M}=\{1,\ldots,M\}$ and the set of edges $\mathcal{E}$ without loops. The weight of an edge $(j,k)$, $j\neq k$, is equal to the Pearson correlation coefficient $W(j,k)$ between features $j$ and $k$ calculated by a given sample.  
\item For any subset of features $\mathcal{S}\subset \mathcal{M}$ we define subgraph $G_{\mathcal{S}}$ of graph $G$.
\item For any subset of features $\mathcal{S}\subset \mathcal{M}$, we define the value of function $v$ as follows:\footnote{The characteristic function \eqref{eq:characteristic-function-graph} was proposed in the paper \cite{suri2008determining}, the authors consider a cooperative game where the characteristic fucntion is defined by the group degree centrality  \cite{everett1999centrality} of each coalition \cite{tarkowski2017game}.}
\begin{equation}
	\label{eq:characteristic-function-graph}
	v(\mathcal{S}) = \sum_{\{j,k\} \subseteq G_{\mathcal{S}}} W(j,k).
\end{equation}

\item We calculate the Shapley value by formula \eqref{shap value}. As a result, the algorithm gives the Shapley value centrality for each node (feature).
\end{enumerate}

Since the set of features $\mathcal{M}$ may be large, we propose an approach for approximating the Shapley value with the high accuracy presented in \cite{zhao2024investigationcentralitymeasuresopinion}, and the proposed method is based on calculation of weights of the edges and function \eqref{eq:characteristic-function-graph}, where $\mathcal{S}$ is a subset of features from set  $\mathcal{M}$, and $G_{\mathcal{S}}$ is the subgraph induced by $\mathcal{S}$.  With a very minor modification, we can apply this approach to define a new XAI method presented in the next section.

\subsubsection{Description and algorithm of the ShapG method}

The ShapG method can be divided into following steps:

\begin{enumerate}
  \item We define an undirected weighted graph $G=(\mathcal{M},\mathcal{E})$, where $\mathcal{M}$ is the set of nodes which are associated with the features $\mathcal{M}=\{1,\ldots,M\}$ in the prediction model and $\mathcal{E}$ is the set of all possible edges without loops, i.e. $\mathcal{E}=\{(i,j): i\in \mathcal{M}, j\in \mathcal{M}, i\neq j\}$ is a complete graph without loops. The weight of an edge $(j,k)$, $j\neq k$, is equal to the Pearson correlation coefficient $W(j,k)$ between features $j$ and $k$ calculated on a given sample.  
  
  \item The matrix of weights $W=\{W(j,k)\}_{(j,k)\in \mathcal{E}}$ is usually a very dense matrix, therefore, we need to reduce the density of graph $G$ to reduce the number of further calculations. We implement the idea of  keeping all features of the dataset while minimizing the number of edges in the graph to reduce the density. The corresponding method is realized in Algorithm~\ref{alg:data-preprocessing}. The idea is straightforward: we construct graph $G'$ starting from the empty graph by iteratively selecting the edges with largest Pearson correlation coefficients given in  matrix $W$, and adding these edges into graph $G'$ ensuring each node is included in $G'$ at least once and graph $G'$ is a connected graph (the latter condition is a stopping rule in Algorithm ~\ref{alg:data-preprocessing}). The output of  Algorithm~\ref{alg:data-preprocessing} is a new graph $G'$. In the following steps, we do not use matrix of  weights $W$.\footnote{We only use weights to reduce graph $G$ to $G'$, we do not use it to calculate the Shapley value.}
  
\item In graph $G'$, we define subgraph $G'_{\mathcal{S}}$ for any subset of features $\mathcal{S}\subset \mathcal{M}$.
\item 
We define characteristic function $f(\mathcal{S})$ assigning the $R^2$ score (for regression models) or $F1$ score (for classification models) for any subset of features $\mathcal{S}$:
\begin{equation}
	\label{eq:characteristic-function-xai}
	v(\mathcal{S}) = f(\mathcal{S}).
\end{equation}
The characteristic function used for XAI can be defined by~\eqref{eq:characteristic-function-xai}, where $f(\mathcal{S})$ is calculated as a prediction given by a ML model trained by only features from subset $\mathcal{S}$. For our purpose, we can use $R^2$ or $F1$ score as a measure of ``power'' of subset $\mathcal{S}$.

\item We calculate the Shapley value by formula \eqref{shap value}. We use Algorithm \ref{alg:shapley-value} to find values of the Shapley value using exact formula \eqref{shap value}. If the number of features is large, we use Algorithm \ref{alg:shapley-approximation} to find the approximated Shapley value.
\end{enumerate}

\begin{algorithm}[h!]
  \caption{Data preprocessing for ShapG method}
  \label{alg:data-preprocessing}
  \begin{algorithmic}[1] 
    \Require Dataset with $M$ features
    \Ensure The adjacency matrix $A\in\mathbb{R}^{M\times M}$.
    \State Compute the Pearson correlation matrix $W$ for all features. \Comment{This is an initial matrix of weights for graph $G$}
    \State Initialize the adjacency matrix $A$ with zero matrix.
    \State $\mathcal{E} \leftarrow \text{List of tuples } (i, j, W(i, j))$, where $i < j$.
    \State $\mathcal{E} \leftarrow$ Sort $\mathcal{E}$ by $|W(i, j)|$ in descending order.
    \State $\mathcal{C} \leftarrow$ Initialize an empty set to store connected nodes.
    \State Initialize index $k = 0$
    \While{$|\mathcal{C}| < M$ and $k < |\mathcal{E}|$ and $G'$ is not connected} \Comment{$G'$ represented by $A$}
        \State $i, j, weight = \mathcal{E}[k]$
        \If{$i \notin \mathcal{C} \lor j \notin \mathcal{C}$}
            \State $A(i, j) \leftarrow 1$
            \State $A(j, i) \leftarrow 1$
            \State $\mathcal{C} \leftarrow \mathcal{C} \cup \{i, j\}$
        \EndIf
        \State $k \leftarrow k + 1$
    \EndWhile
    
    \Return The adjacency matrix $A$ to represent the feature graph.
  \end{algorithmic}
\end{algorithm}

The Algorithm~\ref{alg:shapley-value} describes the calculation of the Shapley value of features $1,\ldots,M$ based on characteristic function defined by~\eqref{eq:characteristic-function-xai} following the steps described above.\footnote{For this algorithm, it's not necessary to reduce the graph density, because regardless of the structure of the graph, Algorithm 2 always needs to traverse all possible coalitions. Algorithm 2 needs to iterate for all possible coalitions, so for graph $G$ and graph $G'$, the number of iterations is the same. For Algorithm 3, we do not need to consider all the coalitions, but only any node and its neighbors. Therefore, Algorithm 2 is graph independent, but Algorithm 3 is graph dependent.}

\begin{algorithm}[h!]
  \caption{Calculation of the Shapley Value based on graph $G'$}
  \label{alg:shapley-value}
  \begin{algorithmic}[1] 
    \Require A graph $G'(\mathcal{M},\mathcal{E})$ with $M = |\mathcal{M}|$ nodes
  \Ensure Shapley value component $\phi(i)$ for each node $i \in \mathcal{M}$
  
  \ForAll{nodes $i \in \mathcal{M}$}
      \State Initialize $\phi(i) \leftarrow 0$
  \EndFor
  
    \ForAll{nodes $i \in \mathcal{M}$}
        \ForAll{subsets $\mathcal{S} \subseteq \mathcal{M} \setminus \{i\}$}
            \State Compute $v(\mathcal{S}) \leftarrow f(\mathcal{S})$
            \State Compute $v(\mathcal{S} \cup \{i\}) \leftarrow f(\mathcal{S} \cup \{i\})$
            \State $\Delta v(\mathcal{S}, i) \leftarrow v(\mathcal{S} \cup \{i\}) - v(\mathcal{S})$
            \State coeff $\leftarrow \frac{S! \cdot (M - S - 1)!}{M!}$
            \State $\phi(i) \leftarrow \phi(i) + \text{coeff} \cdot \Delta v(\mathcal{S}, i)$
        \EndFor
    \EndFor
  
  \Return $\phi(i)$ for all $i \in \mathcal{M}$
  \end{algorithmic}
  \end{algorithm}

However, the Shapley value is pretty computationally expensive,  especially, for a large number of features and consequently large number of feature subsets. 
Therefore, we provide a modified algorithm based on the fact that an influence on a particular node from other nodes is decreasing with an increase of the path length connecting them. Algorithm~\ref{alg:shapley-approximation} can be used instead of Algorithm~\ref{alg:shapley-value} having $G'$ as an input. Algorithm~\ref{alg:shapley-approximation} implements the following ideas to speed up the calculation of the Shapley value:
\begin{enumerate}
  \item[1.] \textit{Depth limitation}: We limit the depth of the set of neighbors considered, the number of subsets used in calculations is reduced.
  
  We set parameter $d_{max}$ which is the depth in the graph for any node $i\in\mathcal{M}$ to form the set of neighbors. Define $\psi(i,d_{max})$ as the set of local neighbors of node $i$ up to depth $d_{max}$ excluding node $i$. Then we calculate the Shapley value of a node/feature $i$ based on the set of local neighbors by equation  \eqref{eq:shapley-value-approximation}, where $\beta=\frac{\psi(i,d_{max})+1}{m+1}$ is the scaling factor, and $m$ is the maximal size of the neighborhood set. 
  
  \item[2.] \textit{Local subset iteration}: We make iterations over subsets only within the local neighbors, rather than within the entire graph. To be specific, we calculate only the marginal contribution of the current node to its neighboring nodes from set $\psi(i,d_{max})$, where $d_{max}$ is set (see Item 1).

  In Algorithm~\ref{alg:shapley-approximation}, we  use formula~\eqref{eq:characteristic-function-xai} to define the values of the characteristic function, but we limit the depth of the neighborhood considered (Item 2) to reduce the number of calculations. 
  
  \item[3.] \textit{Neighborhood size sampling}: For the large number of neighbors, computational complexity is reduced by random sampling of neighbors from the neighborhood. 
  
When $|\psi(i,d_{max})|\geq m$, we choose a random sample of $m$ neighbors from the set $\psi(i,d_{max})$ several times and calculate the Shapley value based on these samples. The number of samples $H_{|\psi(i,d_{max})|,m}$ is given by
\begin{equation}
	\label{eq:sampling-times}
	H_{|\psi(i,d_{max})|,m} = \left(\frac{|\psi(i,d_{max}|)+\frac{1}{2}}{m} - \frac{1}{2}\right)\left(\ln |\psi(i,d_{max})|+\gamma\right)+\frac{1}{2},
\end{equation}
where $\gamma \approx 0.5772156649$ is the Euler-Mascheroni constant. The value $H_{|\psi(i,d_{max})|,m}$ is the mathematical expectation of the number of samples each time collecting $m$ neighbors from the set $\psi(i,d_{max})$ until all neighbors are collected.\footnote{This probabilistic scenario is known as a generalized coupon collector's problem, which was introduced and examined in \cite{PlyaEineWI}.}  As shown in  Algorithm~\ref{alg:shapley-approximation}, we repeat sampling process $H_{|\psi(i,d_{max})|,m}$ times and then take the average value of the Shapley values.
\end{enumerate}

The approximated Shapley value is calculated as follows:
 \begin{equation}
	\label{eq:shapley-value-approximation}
	\phi_a(i) = 
	\begin{cases}
		\sum\limits_{S\subseteq \psi(i,d_{max})}\frac{v(S\cup \{i\})-v(S)}{2^{|\psi(i,d_{max})|}}  &\text{ if } |\psi(i,d_{max})| < m, \\[10pt]
		\beta\sum\limits_{S\subseteq \psi(i,d_{max})}\frac{v(S\cup \{i\})-v(S)}{2^{|\psi(i,d_{max})|}}  &\text{ if } |\psi(i,d_{max})| \geq m.
	\end{cases}
\end{equation}
Items 1--3 from this list reduce the number of calculations and the computational complexity of an algorithm. Based on these three items, the modified Algorithm \ref{alg:shapley-approximation} of the ShapG\footnote{The corresponding code for ShapG can be found in the GitHub repository at \url{https://github.com/vectorsss/shapG}.} method is presented.

  \begin{algorithm}[!h]
    \caption{Approximated Shapley Value based on  graph of features}
    \label{alg:shapley-approximation}
    \begin{algorithmic}[1]
      \Require A graph $G' = (\mathcal{M}, \mathcal{E})$, depth limit $d_{max}$, maximal neighbors size $m$
      \Ensure Shapley value $\phi_a(i)$ for each node $i \in \mathcal{M}$
      
      \State Initialize $\phi(i)_a \gets 0$ for each $i \in \mathcal{M}$
      
      \For{$i \in \mathcal{M}$}
      \State $\psi(i,d_{max}) \gets$ Calculate or retrieve all $neighbors$ of $i$ up to depth $d_{max}$
          \If{$|\psi(i,d_{max})| < m$}
            \For{each subset $\mathcal{S} \subseteq \psi(i,d_{max}) \setminus \{i\}$}
                  \State Compute $v(\mathcal{S}) \leftarrow f(\mathcal{S})$
                  \State Compute $v(\mathcal{S} \cup \{i\}) \leftarrow f(\mathcal{S} \cup \{i\})$
                  \State $\Delta v(\mathcal{S}, i) \leftarrow v(\mathcal{S} \cup \{i\}) - v(\mathcal{S})$
                  \State $\phi_a(i) \gets \phi_a(i) + \Delta v(\mathcal{S}, i)$
              \EndFor
              \State coeff $\leftarrow \frac{1}{2^{|\psi(i,d_{max})|}}$
              \State $\phi_a(i) \gets \phi_a(i) \cdot \text{coeff}$, normalize $\phi_a(i)$ based on the number of subsets
          \Else
          \State Pick up $m$ neighbors randomly from $\psi(i,d_{max})$ and repeat $H_{|\psi(i,d_{max})|,m}$ times
              \For{ i = 1 to $H_{|\psi(i,d_{max})|,m}$}
                  \State $s\_neighbors \gets $ Randomly select a sample of $m$ neighbors from $\psi(i,d_{max})$, 
                  \For{each subset $\mathcal{S} \subseteq s\_neighbors \setminus \{i\}$}
                      \State Calculate $v(\mathcal{S})$ and $v(\mathcal{S} \cup \{i\})$ as before
                      \State $\Delta v(\mathcal{S}, i) \leftarrow v(\mathcal{S} \cup \{i\}) - v(\mathcal{S})$
                      \State $\phi_a(i) \gets \phi_a(i) + \Delta v(\mathcal{S}, i)$
                  \EndFor
              \EndFor
                \State coeff $\leftarrow {1}/{2^{|\psi(i,d_{max})|}}/H_{|\psi(i,d_{max}|),m} \cdot \frac{|\psi(i,d_{max})|+1}{m+1}$
                \State $\phi_a(i) \gets \phi_a(i) \cdot \text{coeff}$
          \EndIf
      \EndFor
      \State \Return $\phi_a(i)$ for all $i \in \mathcal{M}$
    \end{algorithmic}
    \end{algorithm}

\section{Experiments}\label{sec4}
\subsection{Description of datasets}

To demonstrate the work of our XAI method ShapG, we consider two datasets to construct the prediction models: (i) the ``housing price'' dataset for regression prediction, and (ii) the ``H1N1 flu vaccine'' dataset for classification prediction.

We briefly describe datasets:
\begin{enumerate}
\item The ``housing price'' dataset was collected by the U.S. Census Bureau for housing information in the Boston, Massachusetts area. The dataset contains 13 features including ``per capita crime rate by town'', ``average number of rooms per dwelling'', and ``lower status of the population'', as well as a target variable, that is, the ``median value of owner-occupied homes.''

\item The ``H1N1 flu vaccine'' dataset  is provided by the National Center for Health Statistics and borrowed from the DrivenData website. The dataset contains 35 features including an ``individual's age'', ``gender'', ``education level'', and ``knowledge of the H1N1 flu vaccine'', as well as a target variable, that is, a binary value of whether or not an individual received the H1N1 flu vaccine.
\end{enumerate}

These two datasets are used for different prediction tasks, but they are both modeled and predicted by machine learning algorithms. For the regression task, we use $R^2$ to define the characteristic function \eqref{eq:characteristic-function-xai} in our ShapG XAI method, while for classification task, we use $F1$ score to define the values of characteristic function \eqref{eq:characteristic-function-xai}.

We should mention that the ``housing price'' dataset has much less number of features, than the ``H1N1 flu vaccine'' dataset. We will compare our ShapG method with other XAI methods not only quantitatively comparing their explanations but also comparing  running time. Such a comparison helps to evaluate  efficiency of different XAI methods in processing large-scale data.

\subsection{Preprocessing data for ShapG}

\begin{figure}[!h]
  \begin{subfigure}{0.5\textwidth}
    \centering
    \includegraphics[width=\linewidth]{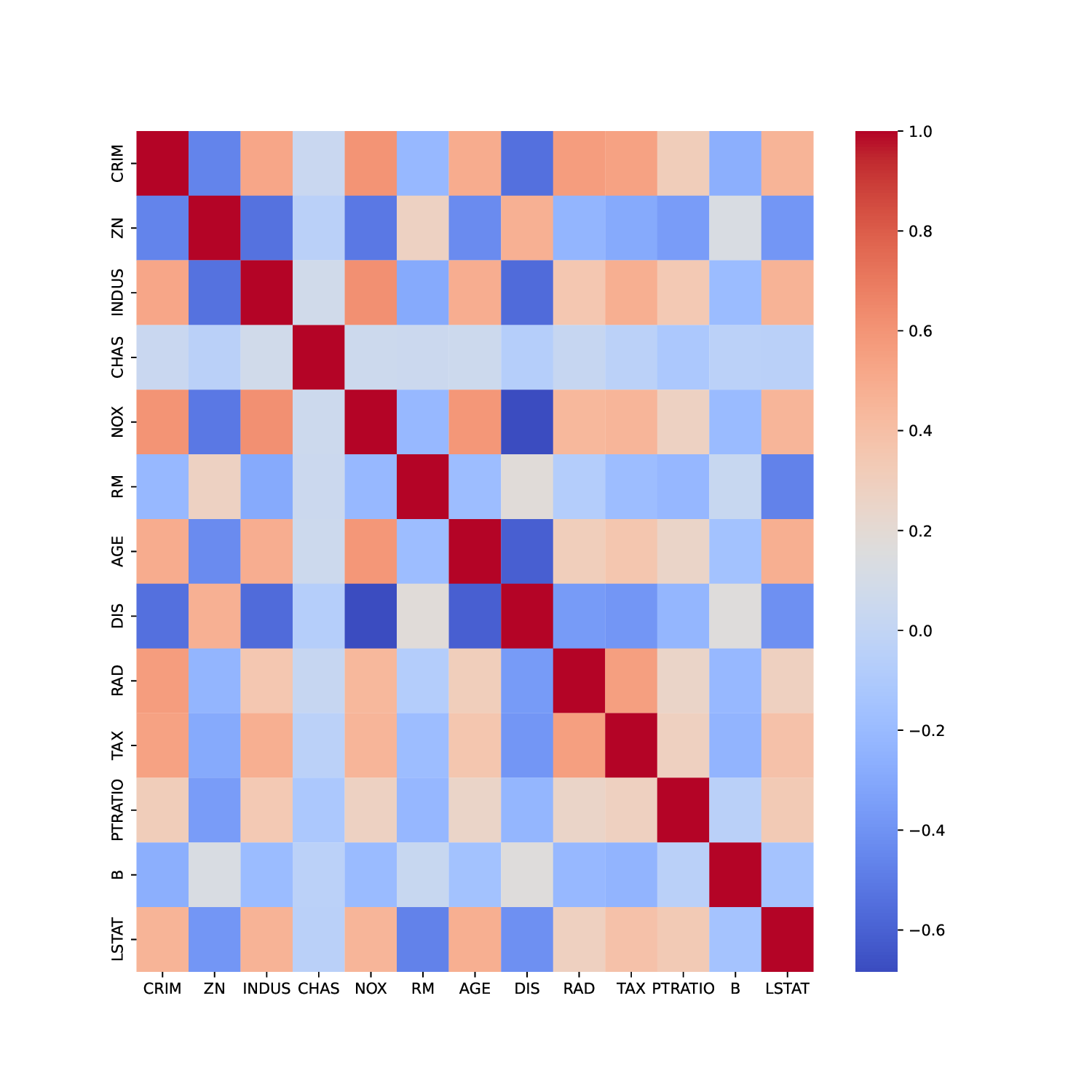}
    \caption{graph $G$ with matrix of weights $W$}\label{fig1_a}
  \end{subfigure}%
  \begin{subfigure}{0.5\textwidth}
    \centering
    \includegraphics[width=\linewidth]{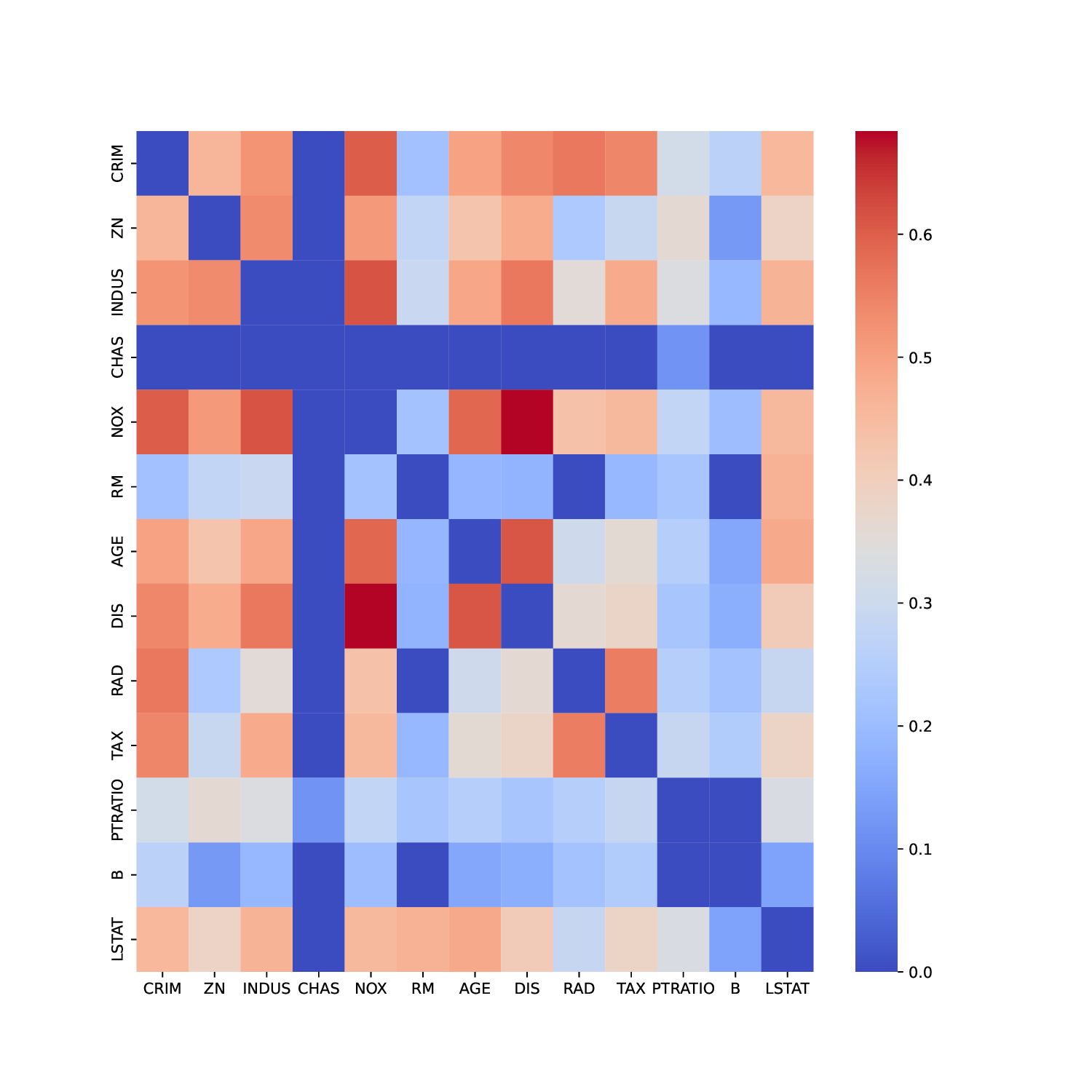}
    \caption{graph $G'$ with matrix of weights $W$(see Algorithm \ref{alg:data-preprocessing})}\label{fig1_b}
  \end{subfigure}
  \caption{Heatmap of Pearson correlation coefficients for the "housing price" dataset}\label{fig1}
\end{figure}

\begin{figure}[!h]
  \begin{subfigure}{0.5\textwidth}
    \centering
    \includegraphics[width=\linewidth]{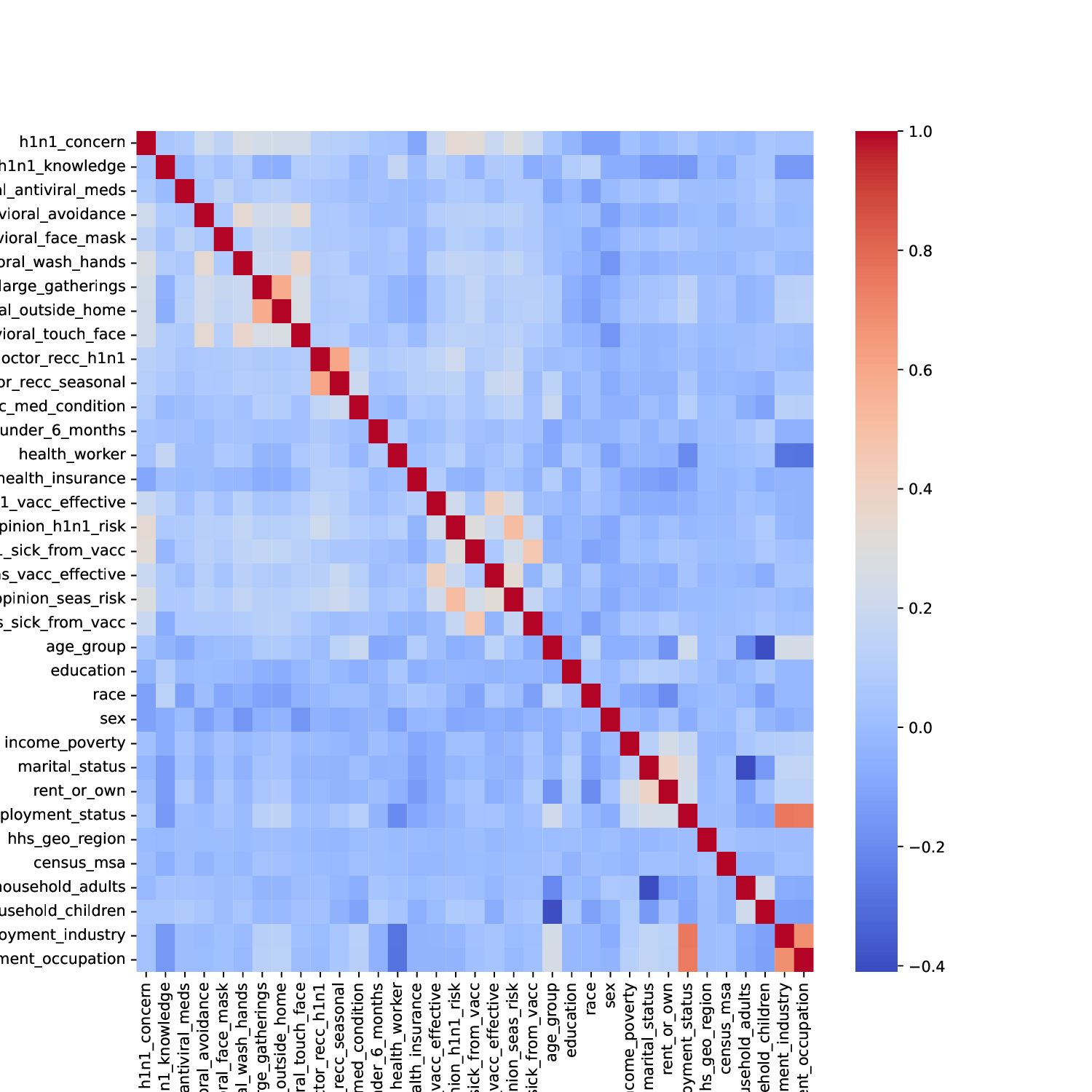}
    \caption{graph $G$ with matrix of weights $W$}\label{fig1_a}
  \end{subfigure}%
  \begin{subfigure}{0.5\textwidth}
    \centering
    \includegraphics[width=\linewidth]{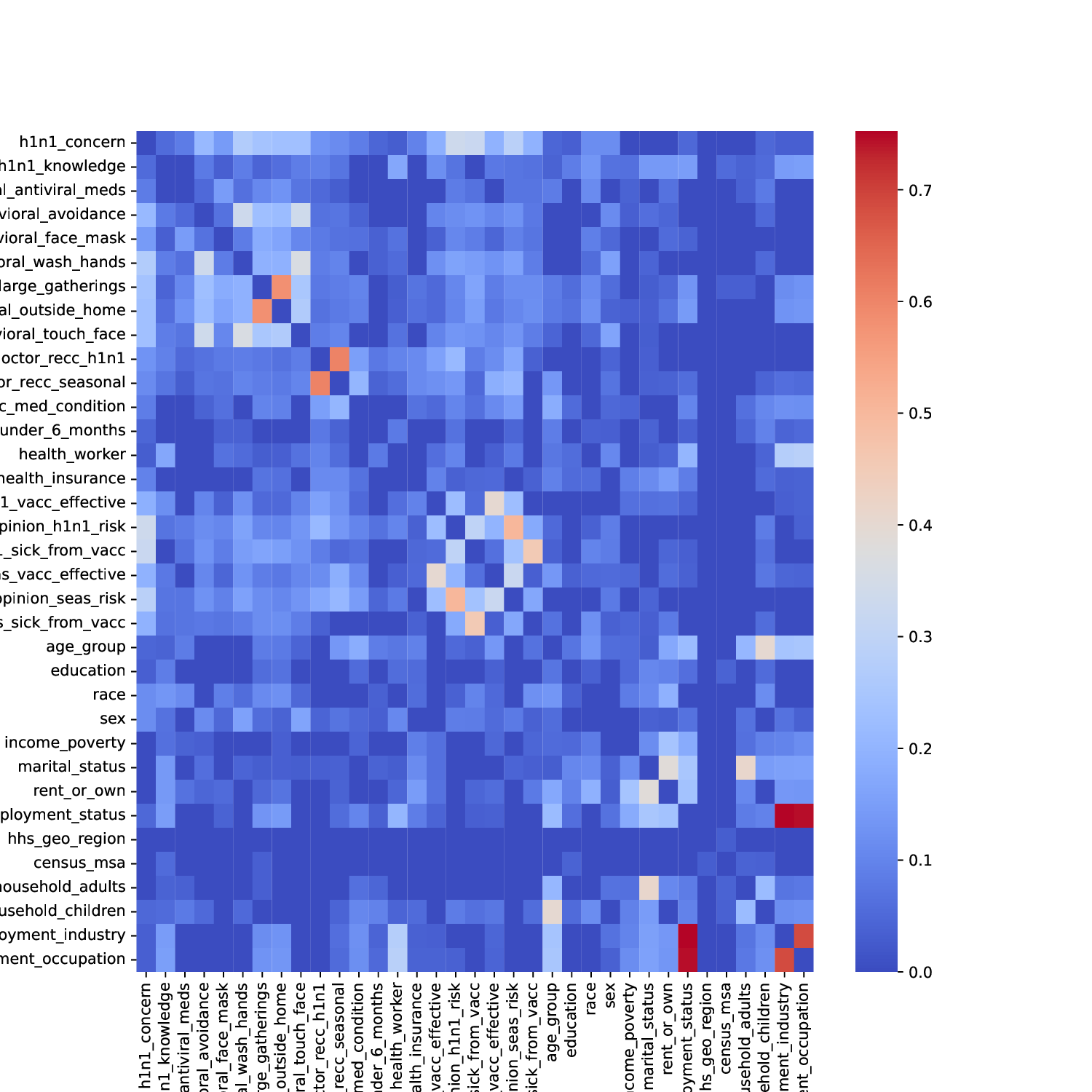}
    \caption{graph $G'$ with matrix of weights $W$(see Algorithm \ref{alg:data-preprocessing})}\label{fig2_b}
  \end{subfigure}
  \caption{Heatmap of Pearson correlation coefficients for the ``H1N1'' dataset}\label{fig2}
\end{figure}

We follow Algorithm \ref{alg:data-preprocessing} to create graph $G'$ based on the original complete graph connecting nodes representing features. Algorithm \ref{alg:data-preprocessing} starts from the empty graph and consequently adds the edges, these are the pairs of features with the strongest correlation coefficients. It stops when all features are connected (graph $G'$ should be connected), ensuring that the feature graph has important structural information. Figures \ref{fig1} and \ref{fig2} represent the feature correlation coefficients heatmaps for the ``housing price'' and ``H1N1'' datasets, respectively. On both figures,  figure (a) shows the original correlation coefficient heatmap, and figure (b) --- the heatmap of correlation coefficients between features in the reduced graph $G'$ computed by Algorithm \ref{alg:data-preprocessing}. 

We highlight that we use Algorithm \ref{alg:data-preprocessing} to reduce the number of edges while preserving all features, thus ignoring ``unimportant'' relationships between features when constructing graph $G'$. This will reduce the number of iterations to compute the components of the Shapley value measuring importance of the features.

\begin{figure}[!h]
  \begin{subfigure}[normla]{.45\linewidth}
  	\centering
	\includegraphics[scale=0.5]{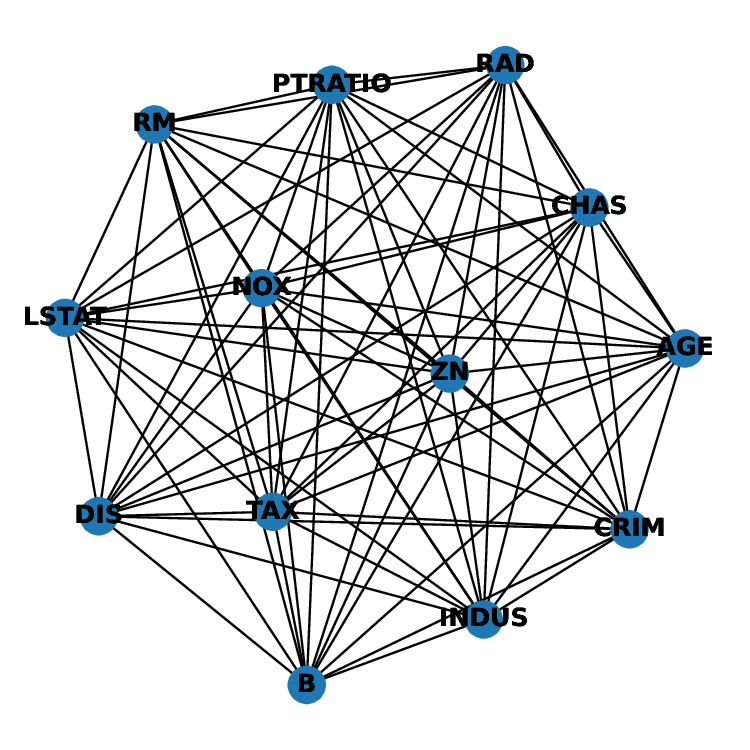}
	\subcaption{Original graph $G$}
	\label{fig3}
\end{subfigure}
\begin{subfigure}[normla]{.45\linewidth}
	\centering
	\includegraphics[scale=0.5]{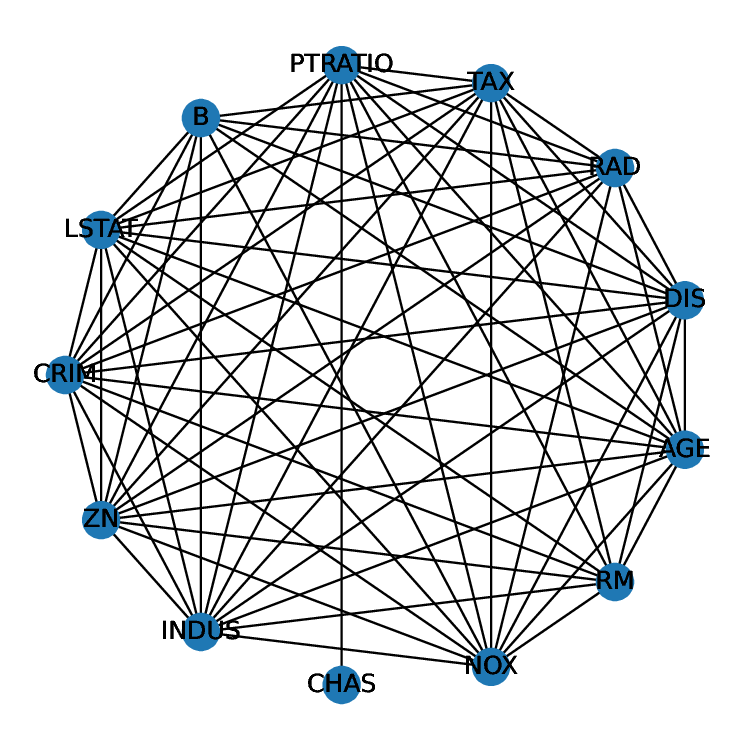}
	\subcaption{Reduced graph $G'$ computed by Algorithm \ref{alg:data-preprocessing}}
	\label{fig4}
\end{subfigure}
\caption{Graph connecting features in ``housing price'' dataset}
\label{fig:graphs housing}	
\end{figure}

\begin{figure}[!h]
	\begin{subfigure}[normla]{.45\linewidth}
		\centering
		\includegraphics[scale=0.15]{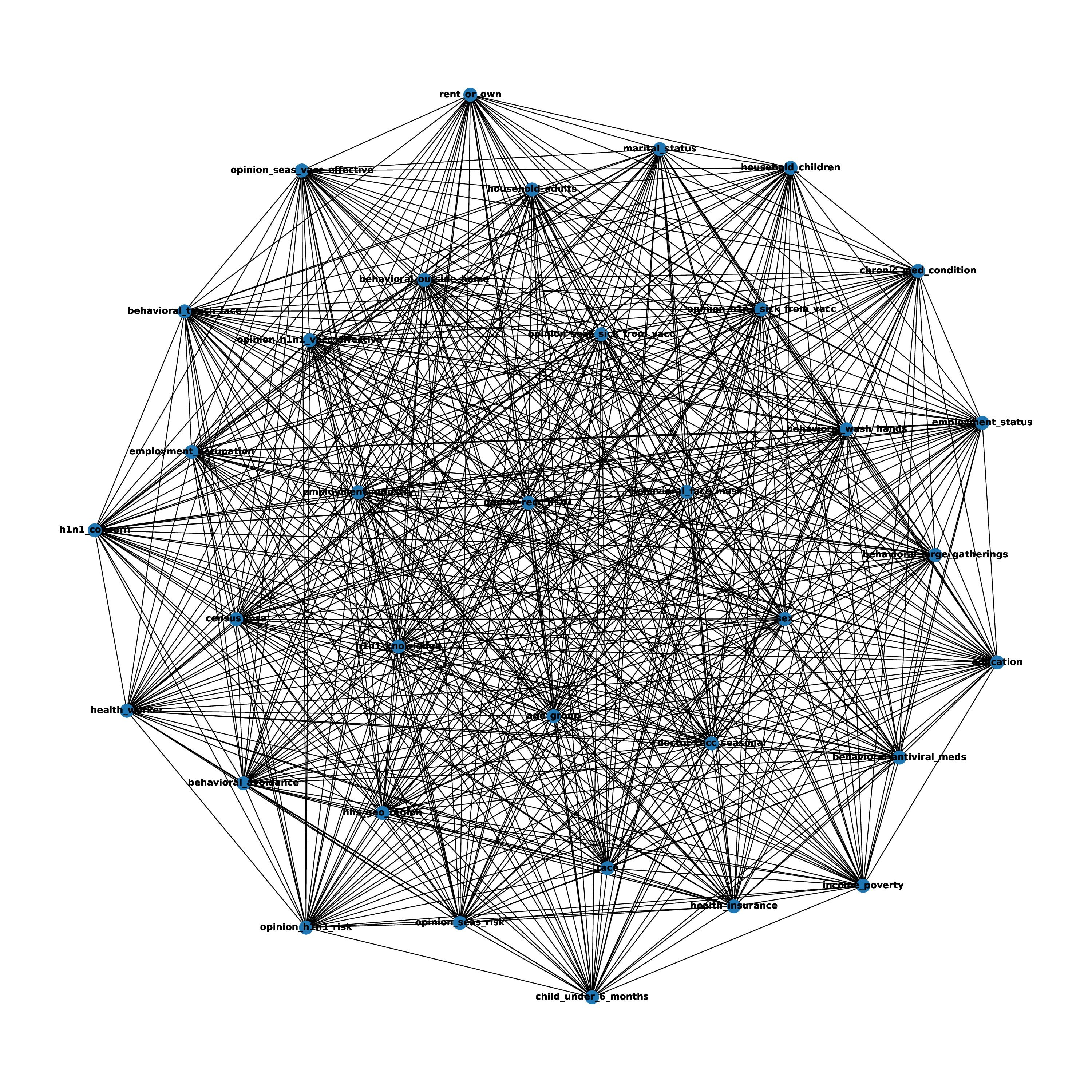}
		\subcaption{Original graph $G$}
		\label{fig5}
	\end{subfigure}
	\begin{subfigure}[normla]{.45\linewidth}
		\centering
		\includegraphics[scale=0.15]{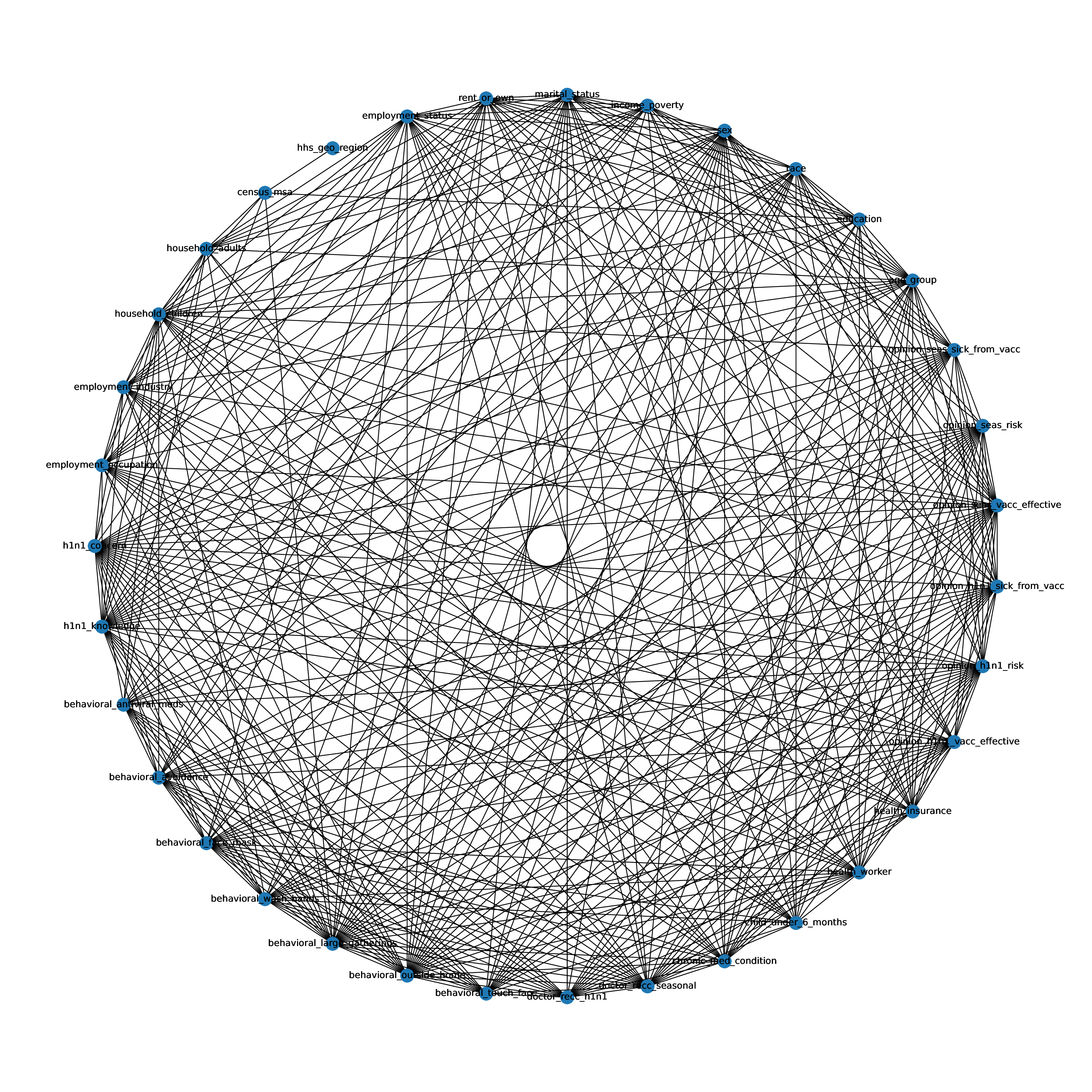}
		\subcaption{Reduced graph $G'$ computed by Algorithm \ref{alg:data-preprocessing}}
		\label{fig6}
	\end{subfigure}
	\caption{Graph connecting features in ``H1N1'' dataset}
	\label{fig:H1N1}	
\end{figure}

The original undirected graph $G$ with the nodes representing features of the ``housing price'' (``H1N1'') dataset is given in Figure \ref{fig3} (Figure \ref{fig5}), while Figure \ref{fig4} (Figure \ref{fig6}) shows the undirected graph $G'$ calculated by Algorithm \ref{alg:data-preprocessing}. So, we simplify graph structures and keep important feature pairs improving running efficiency and explainability of the ShapG method.

\subsection{AI prediction models}\label{sec:models}

To evaluate efficiency of our method ShapG, we apply several Explainable Artificial Intelligence (XAI) methods including our method to explain the importance of features within LightGBM and MLP (Multilayer Perceptron) models.  Moreover, our ShapG algorithm can provide explanations for complex AI models that existing XAI methods cannot explain in a reasonable running time. We adopt ensemble learning and a two-by-two combination of tree models, neural network models, linear models, machine learning models to construct hybrid prediction models. We construct these hybrid models as combination of different types of simpler models to achieve better performance and explainability. There is a list of constructed models (including their combinations) and their descriptions:
\begin{itemize}
  \item \textbf{LightGBM (LGB):} It is an efficient Gradient Boosting Decision Tree (GBDT) framework, which has fast training speed and high performance. It is widely used in practice \cite{ke2017lightgbm}.
  
  \item \textbf{Multilayer Perceptron (MLP):} It is a feed-forward artificial neural network consisting of fully connected neurons with nonlinear activation functions \cite{samatin2016novel}. Through several tests, we adjust the parameters to obtain the best model prediction. In the regression model (for ``housing price'' dataset), a hidden layer containing 300 neurons is used, the learning rate is set to 0.007, the activation function is chosen as ReLU (Rectified Linear Unit), and the optimization algorithm is Adam (Adaptive Moment Estimation). These parameters achieve the best accuracy of prediction for ``housing price'' dataset. In the classification model (for ``H1N1'' dataset), to obtain a high performance, the model has two hidden layers, the first one is with 100 neurons, and the second one is with 50 neurons. The model performs up to 3 iterations to complete training dataset, and the activation function is also chosen as ReLU.
  
  \item \textbf{Ensemble Learning (Stacking):} It is an approach to get better predictive performance by combining several single models \cite{polikar2012ensemble}. Stacking is one of the ensemble learning methods, where first several different types of base models are trained using a training sample, and second, meta model is trained using the predictions of base models as input features in combination with real labels \cite{divina2018stacking}. In this paper, we choose Random Forest and XGBoost as basic models, and LightGBM as the meta model.
  
  \item \textbf{Linear Regression -- LightGBM (Linear - LGB):} It is a classic linear regression model and the more representative LightGBM model for regression prediction on the ``housing price'' dataset. Linear models effectively capture linear relationships between features \cite{weisberg2005applied}, while tree models are able to handle non-linear relationships and high-dimensional features.
  
  \item \textbf{Logistic Regression -- LightGBM (Logistic - LGB):} Logistic regression is a common linear model used for classification problems \cite{lavalley2008logistic}. For ``H1N1'' dataset, we combined logistic regression and LightGBM for classification.
  
  \item \textbf{Linear Regression -- Multilayer Perceptron (Linear - MLP):} Multilayer Perceptron (MLP) is an artificial neural network (ANN) consisting of multiple layers of interconnected neurons with ability to process various data types \cite{taud2018multilayer}. We combined linear regression model and neural network model -- MLP to construct a regression prediction model.
  
  \item \textbf{Logistic Regression -- Multilayer Perceptron  (Logistic - MLP):} We combine Logistic Regression and MLP to make a classification model. Using logistic regression to initially categorize the input data, we then apply MLP to capture more complex patterns and nonlinear relationships. It is suitable for dealing with a variety of complex classification problems.
  
  \item \textbf{Linear Regression -- K-Nearest Neighbors (Linear - KNN):} KNN is a common machine learning algorithm that is widely used to solve classification and regression problems \cite{guo2003knn}. The Linear-KNN model can be used to predict both linear and nonlinear relationships when dealing with regression problems, improving the flexibility and accuracy of the model.
  
  \item \textbf{Logistic Regression -- K-Nearest Neighbors (Logistic - KNN):} We combine logistic regression with KNN algorithm, and this hybrid model is able to utilize both the linear relationship of logistic regression and nonparametric properties of KNN when dealing with classification problems.
  
  \item \textbf{Multilayer Perceptron -- LightGBM (MLP - LGB):} A hybrid model of Multilayer Perceptron (MLP) neural networks and LightGBM can simultaneously utilize the nonlinear fitting ability of neural networks and efficient performance of gradient boosted decision trees.
  
  \item \textbf{K-Nearest Neighbors -- LightGBM (KNN - LGB):} The hybrid model of KNN and LightGBM gives an advantage of the high performance and efficiency of LightGBM while exploiting nonparametric properties of KNN.
  
  \item \textbf{Multilayer Perceptron -- K-Nearest Neighbors  (MLP - KNN):} We combine a multilayer perceptron (MLP) neural network with KNN algorithm, which is capable of exploiting both the nonlinear fitting ability of the neural network and  nonparametric properties of KNN to provide better classification or regression performance.
\end{itemize}

The goal of the hybrid models is to combine different types of models for better performance and explainability. Our proposed XAI algorithm explains these complex AI models and demonstrates its broad applicability. This means that it can be applied to a wide range of complex hybrid models and is not limited by specific model types. Therefore, it is model-agnostic.

We first calculate feature importance for LightGBM and MLP models for two datasets in Section \ref{FI simple}, and then for more complex models presented in the above given list in Section  \ref{FI complex}. 

\subsection{Evaluation of XAI methods}

A common way of evaluating XAI methods is by performing a perturbation analysis of the features, removing features in order of their importance from the largest to the smallest, and observing a decrease of accuracy or $R^2$ of the model~\cite{schlegel2019towards}. 

When we consequently remove features based on their importance, if  performance of a model decreases significantly, it indicates that the feature is very important. By evaluating the XAI method in this way, we can understand the contribution of each feature to the model's predictive performance. 

In order to prove an efficiency of our ShapG method, we can use this evaluation method to compare ShapG with existing popular XAI methods. First, we will apply our proposed XAI approach and other popular XAI methods to generate explanation results. Then, we will remove features gradually according to their importance and observe the changes in the model's performance. We compare this process using the following XAI methods: Feature Importance, Permutation Feature Importance, LIME, SHAP (KernelSHAP and SamplingSHAP), and ShapG.

\section{Results and analysis}\label{sec5}

\subsection{Feature importance calculated by ShapG}\label{FI simple}

\subsubsection{``Housing price'' dataset}

\begin{figure}[!h]
	\begin{subfigure}[b]{.45\linewidth}
		\centering
		\includegraphics[scale=0.4,clip, trim=0cm 0cm 0cm 2.92cm]{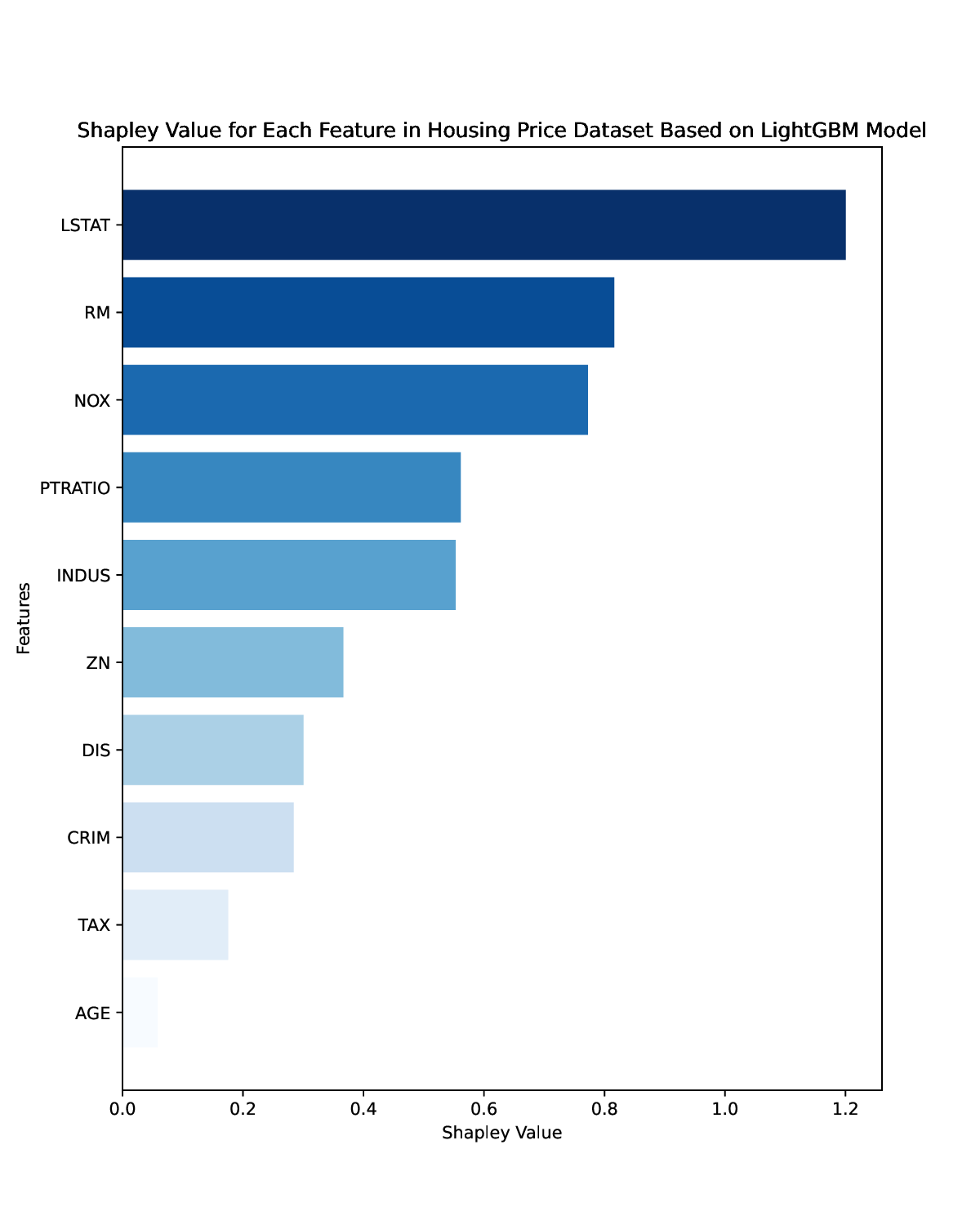}
		\subcaption{LightGBM model}
		\label{fig7}
	\end{subfigure}
\hfill
	\begin{subfigure}[b]{.45\linewidth}
	\centering
	\includegraphics[scale=0.4,clip, trim=0cm 0cm 0cm 2.92cm]{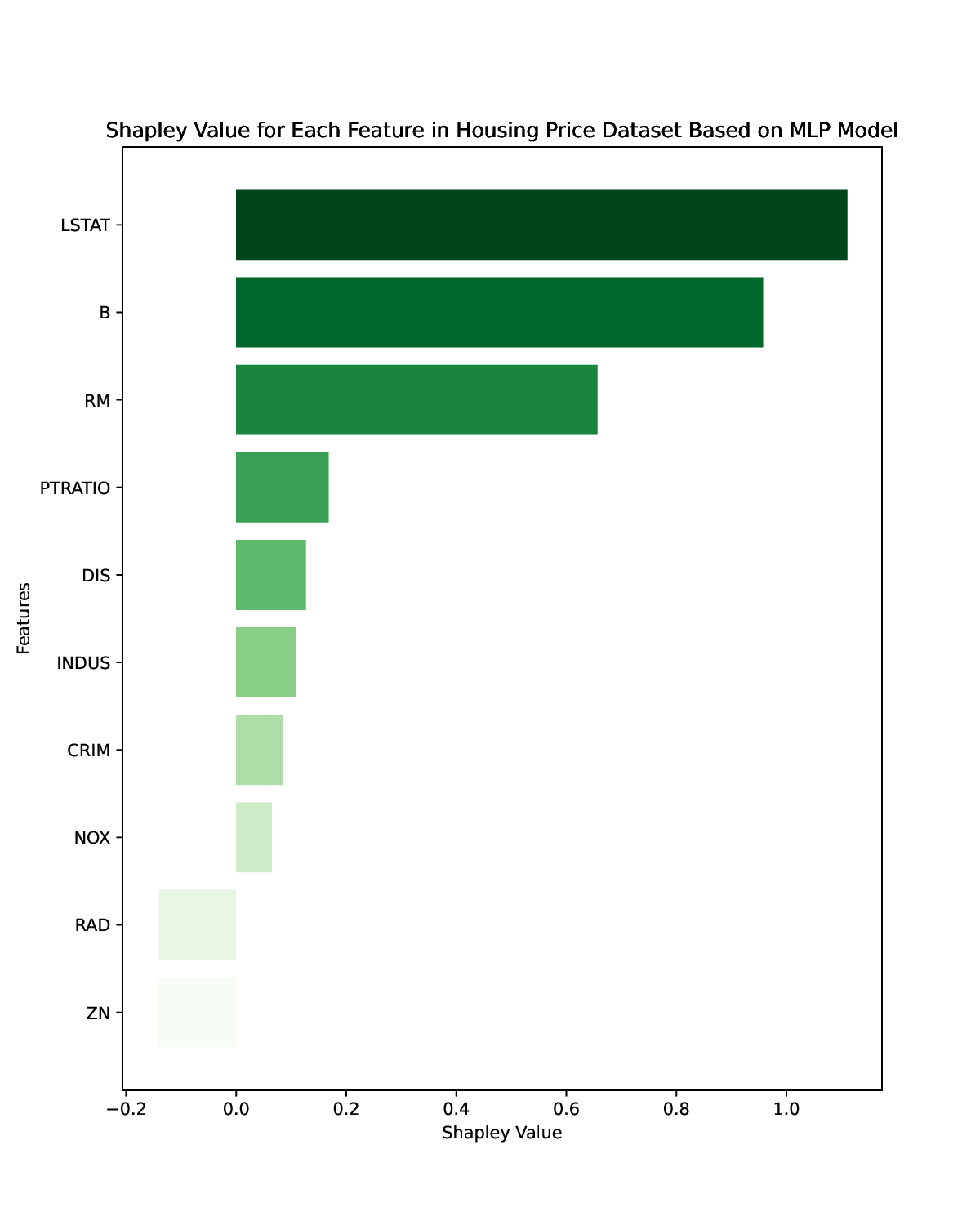}
	\subcaption{MLP model}
	\label{fig9}
\end{subfigure}
	\caption{Feature importance in ``housing price'' dataset calculated with ShapG}
	\label{fig:HP ShapG}	
\end{figure}

Figure \ref{fig:HP ShapG} shows the feature importance calculated by ShapG algorithm for the ``housing price'' dataset based on LightGBM (Fig. \ref{fig7}) and MLP (Fig. \ref{fig9}) models, respectively. For the ``housing price'' dataset with regression prediction based on LightGBM, the four most important features given by ShapG are ``LSTAT (lower status of the population)'', ``RM (average number of rooms per dwelling)'', ``NOX (nitric oxides concentration)'', ``PTRATIO (pupil-teacher ratio by town)'', while with regression prediction based on MLP model, these features are ``LSTAT (lower status of the population)'', ``B (the proportion of blacks by town)'', ``RM (average number of rooms per dwelling)'', ``PTRATIO (pupil-teacher ratio by town)''. The three features are the same for both models, and the most important one is the unique for both models too.

\subsubsection{``H1N1'' dataset}

\begin{figure}[!h]
	\begin{subfigure}[b]{.45\linewidth}
		\centering
		\includegraphics[scale=0.35,clip, trim=0cm 0cm 0cm 0.8cm]{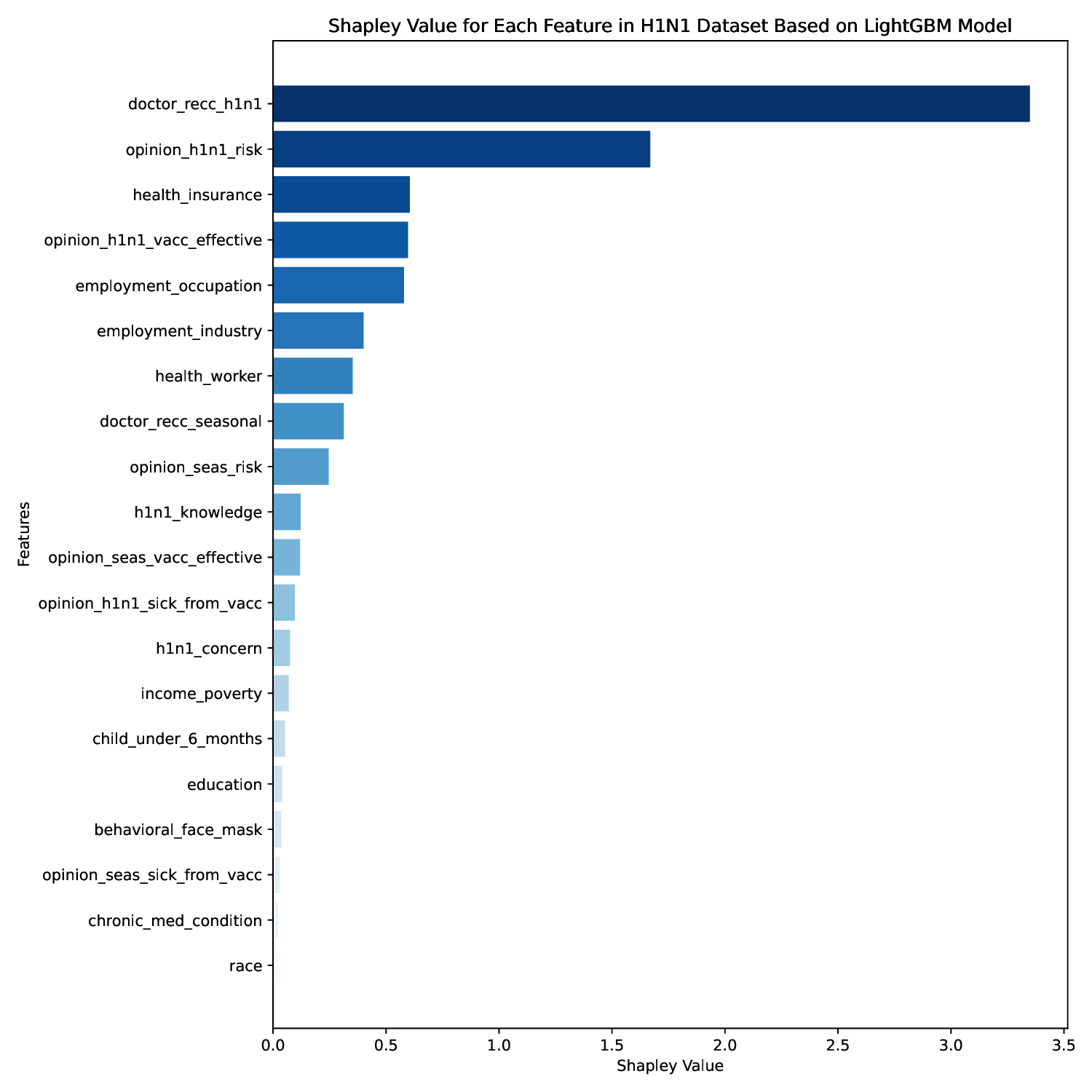}
		\subcaption{LightGBM model}
		\label{fig8}
	\end{subfigure}
	\hfill
	\begin{subfigure}[b]{.45\linewidth}
		\centering
		\includegraphics[scale=0.32,clip, trim=0cm 0cm 0cm 0.8cm]{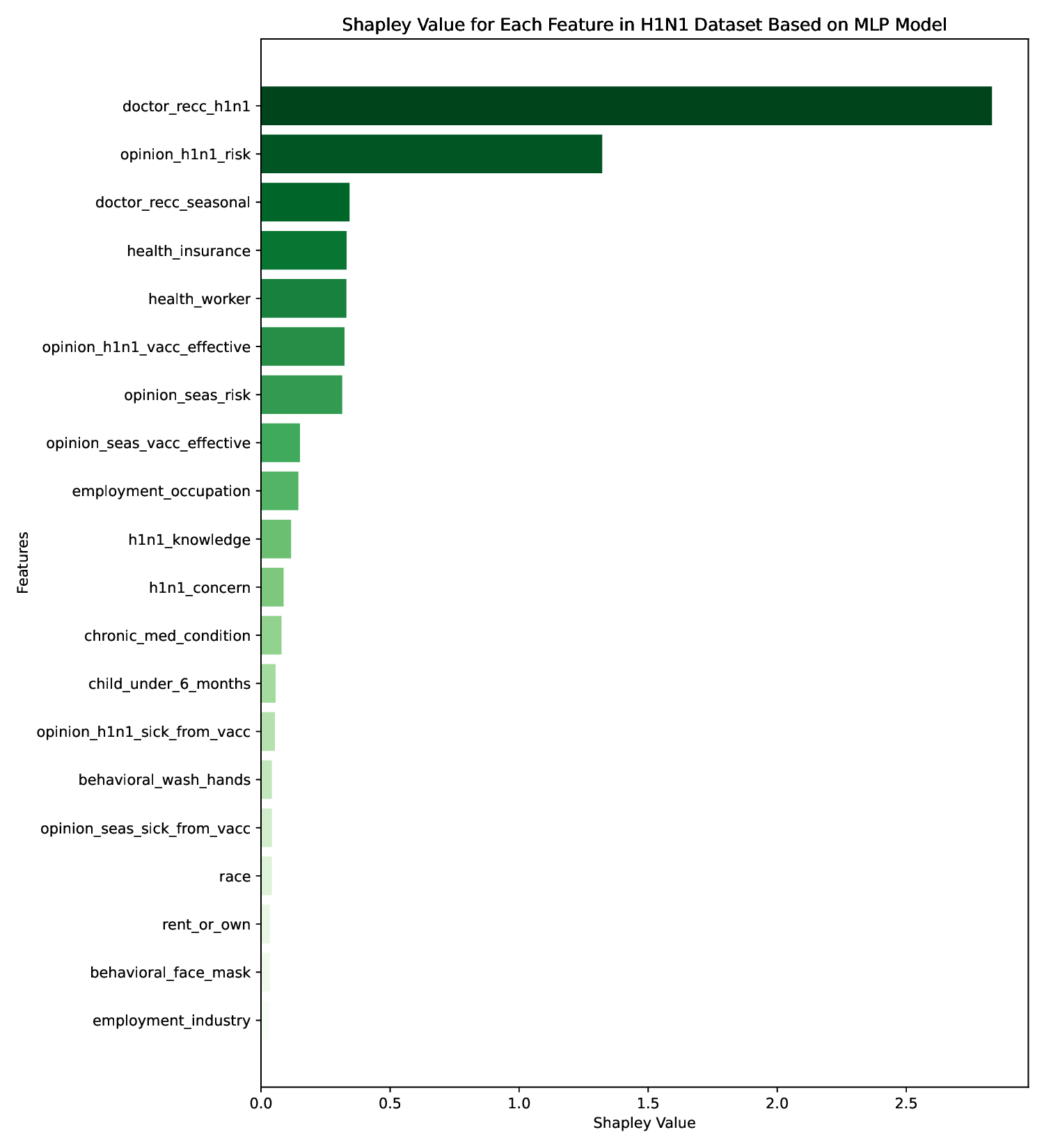}
		\subcaption{MLP model}
		\label{fig10}
	\end{subfigure}
	\caption{Feature importance in ``H1N1'' dataset calculated with ShapG}
	\label{fig:H1N1 ShapG}	
\end{figure}

Figures \ref{fig:H1N1 ShapG} shows the feature importance calculated by ShapG algorithm for the ``H1N1'' dataset based on LightGBM (Fig. \ref{fig8}) and MLP (Fig. \ref{fig10}) models, respectively. For the ``H1N1'' dataset with classification prediction, the feature's approximated Shapley values given by ShapG algorithm represent the importance of each feature for people's willingness to be vaccinated against H1N1. For classification based on LightGBM, the five most important features are ``doctor recc h1n1 (H1N1 flu vaccine was recommended by doctor)'', ``opinion h1n1 risk (Respondent's opinion about risk of getting sick with H1N1 flu without vaccine)'', ``health insurance'', ``opinion h1n1 vacc effective (Respondent's opinion about H1N1 vaccine effectiveness)'', ``employment occupation (Type of occupation of respondent)''. 

For classification based on MLP model, the five most important features are ``doctor recc h1n1 (H1N1 flu vaccine was recommended by doctor)'', ``opinion h1n1 risk (Respondent's opinion about risk of getting sick with H1N1 flu without vaccine)'', ``doctor recc seasonal (Seasonal flu vaccine was recommended by doctor)'', ``health insurance", ``health worker''. Three most important features for both  LightGBM and MLP models coincide.

Based on the XAI approach to explain different AI models, it produces different explanation results since different model architectures may have differences in processing data, extracting features and predictions, this leads to differences in the result explanation. When both AI models (LightGBM and MLP) give the same most important features, we can make better decisions and understand the behavior of ``black-box'' models. As mentioned above, for the ``housing price'' dataset, we can observe that features ``LSTAT'', ``RM'', and ``PTRATIO'' are considered to be the most important features for both LightGBM and MLP models to predict house prices. For ``H1N1'' dataset, such three features are ``doctor recc h1n1'', ``opinion h1n1 risk (Respondent's opinion about risk of getting sick with H1N1 flu without vaccine)'', and  ``health insurance'', these  are considered to be the most important ones to predict if a person is willing to be vaccinated thereby developing relevant strategies to increase vaccination rates.

\subsection{Evaluation of XAI methods}

In order to prove efficiency of our proposed XAI method ShapG, we compare the results of its work with other existing XAI methods by (i) evaluating all methods based on perturbation of features, and (ii) measuring running time to obtain results. 

We introduce the feature numbering for ``housing price'' and ``H1N1'' in Tables \ref{tab1} and \ref{tab2}, respectively.

\begin{table}[!htbp]
	\centering
	\caption{Feature No. in ``house price'' dataset}\label{tab1}%
	\begin{tabular}{|c|c||c|c||c|c|}
		\hline
		\textbf{No.} & \textbf{Feature} & \textbf{No.} & \textbf{Feature} & \textbf{No.} & \textbf{Feature} \\
		\hline
		1 & CRIM & 	6 & RM & 11 & PTRATIO \\
		\hline
		2 &ZN & 7 & AGE & 12 &  B\\
		\hline
		3 & INDUS & 8 & DIS	& 13 & LSTAT\\
		\hline
		4 & CHAS & 9 & RAD & &\\
		\hline
		5 & NOX &  10 & TAX & & \\
		\hline
	\end{tabular}
\end{table}

\begin{table}[!htbp]
	\centering
	\caption{Feature No. in ``H1N1'' dataset}\label{tab2}%
	\begin{tabular}{|c|c||c|c|}
		\hline
		\textbf{No.} & \textbf{Feature} & \textbf{No.} & \textbf{Feature} \\
		\hline
		1 & h1n1 concern & 19 & opinion seas vacc effective\\
		\hline
		2 &h1n1 knowledge & 20 & opinion seas risk\\
		\hline
		3 & behavioral antiviral meds & 21 & opinion seas sick from vacc\\
		\hline
		4 & behavioral avoidance & 22 & age group \\
		\hline
		5 & behavioral face mask & 23 & education\\
		\hline
		6 & behavioral wash hands & 24 & race\\
		\hline
		7 & behavioral large gatherings & 25 & sex  \\
		\hline
		8 & behavioral outside home & 26 & income poverty \\
		\hline
		9 & behavioral touch face & 27 & marital status\\
		\hline
		10 & doctor recc h1n1 & 28 & rent or own \\
		\hline
		11 & doctor recc seasonal & 29 & employment status \\
		\hline
		12 & chronic med condition & 30 & hhs geo region \\
		\hline
		13 & child under 6 months & 31 & census msa \\
		\hline
		14 & health worker & 32 & household adults \\
		\hline
		15 & health insurance & 33 & household children\\
		\hline
		16 & opinion h1n1 vacc effective & 34 & employment industry\\
		\hline
		17 & opinion h1n1 risk & 35 & employment occupation\\
		\hline
		18 & opinion h1n1 sick from vacc & & \\
		\hline
	\end{tabular}
\end{table}

\begin{figure}[!htbp]
    \centering
    \scalebox{0.5}{\includegraphics[clip, trim=0cm 0cm 0cm 2.32cm]{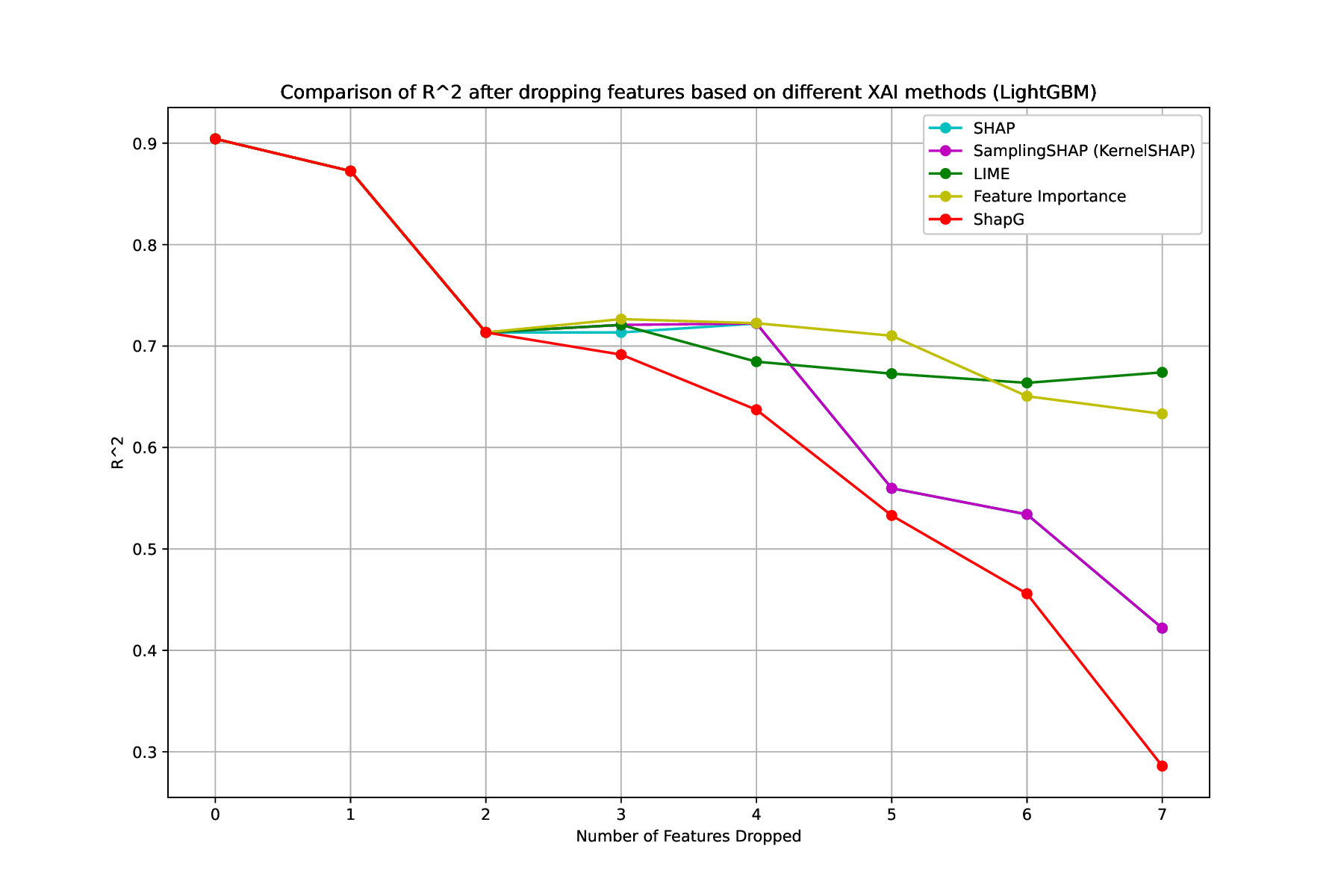}}
    \caption{Comparison of $R^2$ after dropping features based on different XAI methods in ``housing price'' dataset (LightGBM)}\label{fig11}
\end{figure}

\begin{figure}[!htbp]
	\centering
	\scalebox{0.5}{\includegraphics[clip, trim=0cm 0cm 0cm 2.32cm]{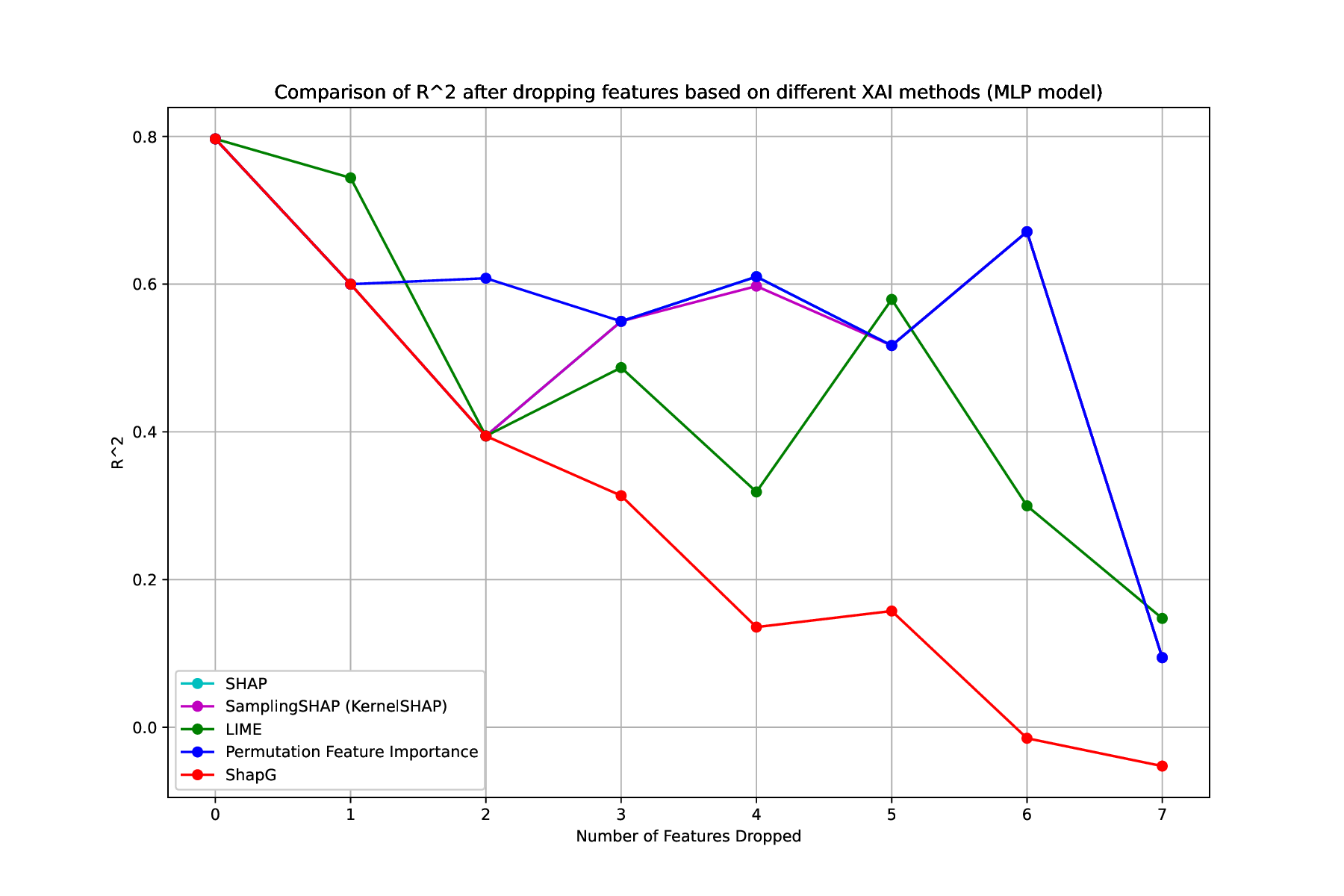}}
	\caption{Comparison of $R^2$ after dropping features based on different XAI methods in ``housing price dataset'' (MLP model)}\label{fig13}
\end{figure}

Table \ref{tab3} shows the results (ranking of feature importance) given by different XAI methods for ``housing price'' dataset based on LightGBM and MLP models. We observe that SamplingSHAP and KernelSHAP give the same results for both LightGBM and MLP models, so they are combined into one column. As we can see in Table \ref{tab3}, for the LightGBM model for the ``house price'' dataset, all XAI methods have the same ranking of the first and second most important features, ``LSTAT (lower status of the population)'' and ``RM (average number of rooms per dwelling)''. But the ranking given by the feature importance method is different. The XAI methods applied for the MLP model including SHAP, SamplingSHAP, KernelSHAP, and ShapG, give the same ranking on the first and second most important features ``LSTAT'' and ``B (the proportion of blacks by town)'', while the ranking given by LIME is the opposite. In addition, the PFI method already disagrees with the results of other XAI methods in the ranking of the second feature. We use the XAI evaluation method to compare performance of these XAI methods, and comparison results are shown in Figures \ref{fig11} and \ref{fig13}.

\begin{table}[h!]
	\centering
	\caption{Feature importance ranking for ``housing price'' dataset}\label{tab3}%
	\vspace{0.2cm}
	\textit{LightGBM model}
	\vspace{0.2cm}
	\begin{tabular}{|c|c|c|c|c|c|}
		\hline
		\diagbox{Rank}{XAI} & SHAP & SamplingSHAP (KernelSHAP) & LIME  &Feature Importance& \textbf{ShapG}\\
		\hline
		Top 1& 13 &13 & 13  &13& 13\\   
		\hline
		Top 2& 6 & 6 & 6 & 6 & 6\\   
		\hline
		Top 3& 8 & 7 & 7 & 1 &5\\   
		\hline
		Top 4& 7 & 8 & 11 & 8 & 11 \\   
		\hline
		Top 5& 5 & 5 & 10 & 7 & 3 \\  
		\hline
		Top 6& 1 & 1 & 12 & 11 & 2\\  
		\hline
		Top 7&  11  & 11 & 4 & 12 & 8\\ \hline 
		\multicolumn{6}{c}{\,} \\
		\multicolumn{6}{c}{\textit{MLP model}}\\
		\hline 
		\diagbox{Rank}{XAI} & SHAP & SamplingSHAP (KernelSHAP) & LIME  & Permutation Feature Importance & \textbf{ShapG}\\
		\hline
		Top 1 &13 & 13 & 12  &13& 13\\   
		\hline
		Top 2& 12 & 12 & 13& 7 & 12\\   
		\hline
		Top 3& 7 & 7 & 11 & 12 &6\\   
		\hline
		Top 4 & 2 & 9 & 3 & 2 & 11 \\   
		\hline
		Top 5 & 9 &2  & 1 & 9 & 8 \\  
		\hline
		Top 6 & 8 & 8& 7 & 8 & 3\\  
		\hline
		Top 7 & 6 & 6 & 6 & 6 & 1\\  
		\hline
	\end{tabular}
\end{table}

\begin{figure}[!htbp]
    \centering
    \scalebox{0.5}{\includegraphics[clip, trim=0cm 0cm 0cm 2.32cm]{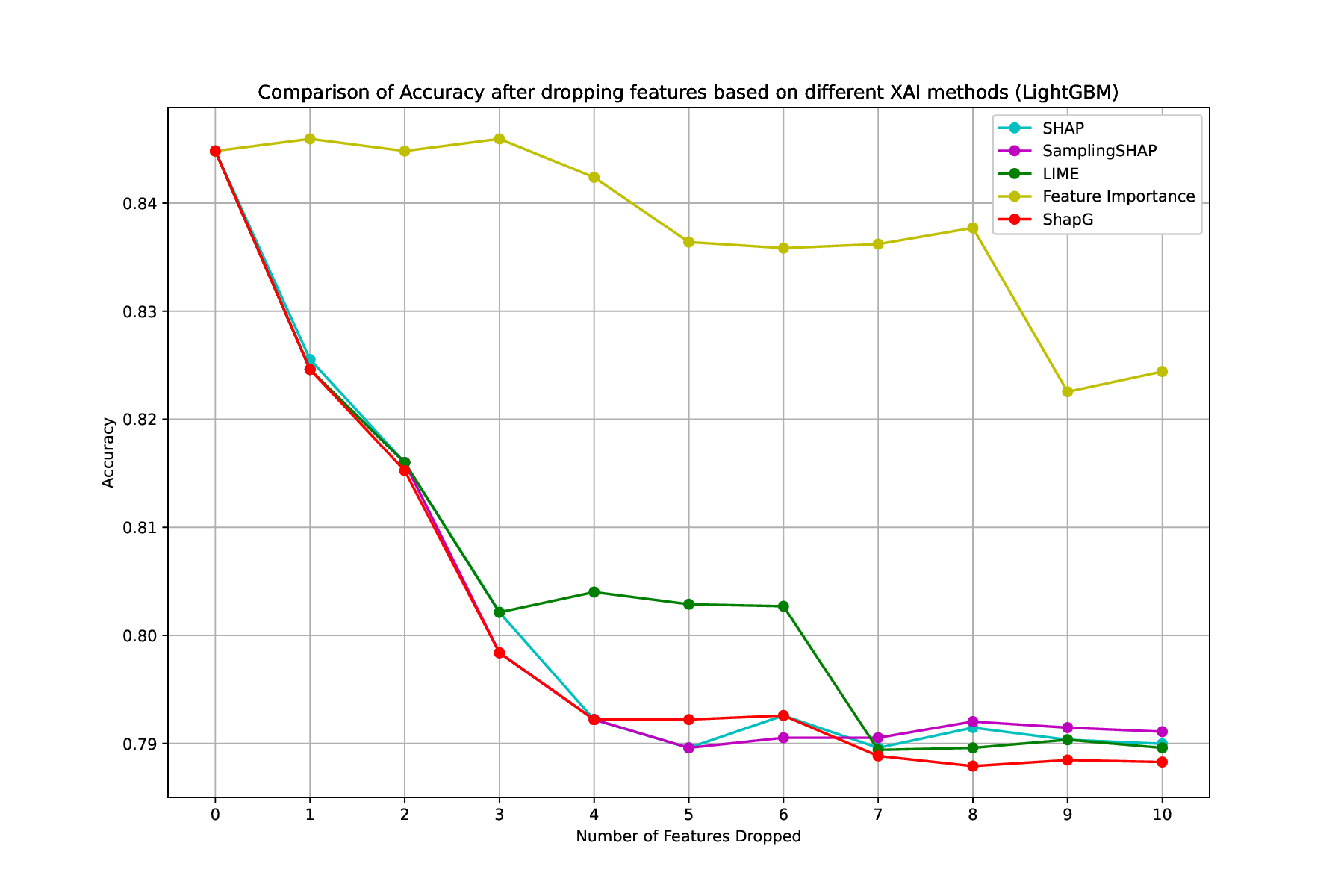}}
    \caption{Comparison of accuracy after dropping features based on different XAI methods in ``H1N1'' dataset (LightGBM)}\label{fig12}
\end{figure}

\begin{figure}[!htbp]
    \centering
    \scalebox{0.5}{\includegraphics[clip, trim=0cm 0cm 0cm 2.32cm]{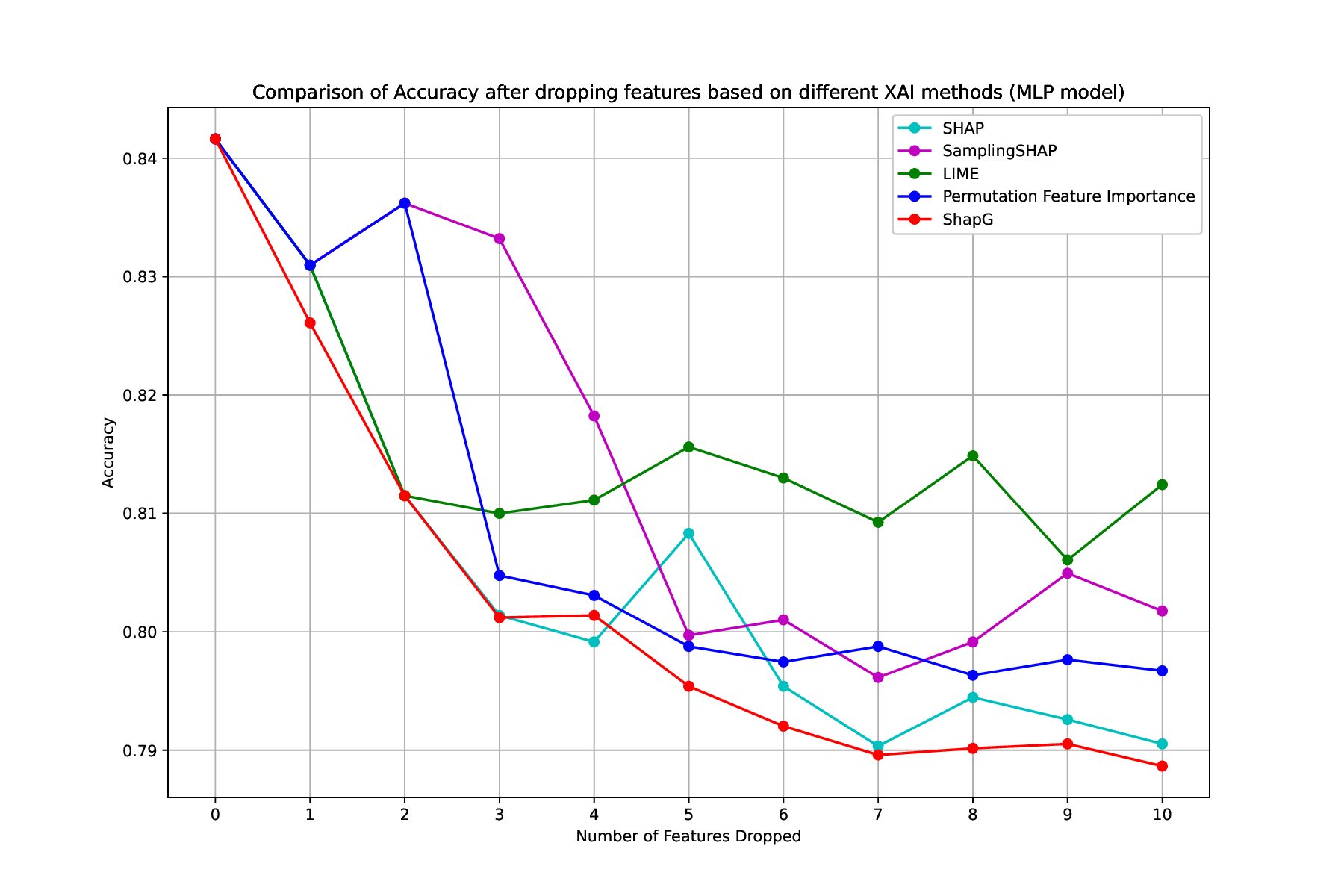}}
    \caption{Comparison of accuracy after dropping features based on different XAI methods in ``H1N1'' dataset (MLP model)}\label{fig14}
\end{figure}

Table \ref{tab4} shows the results (ranking of feature importance) of different XAI methods for LightGBM and MLP models constructed for ``H1N1'' dataset. We should  mention that we do not use KernelSHAP to get explanation results due to its very large running time: it requires more than 72 hours for LightGBM and more than 654 hours for MLP model. As we can see in Table \ref{tab4}, SamplingSHAP, LIME, and ShapG give the same ranking for the most important feature ``doctor recc h1n1 (10)'' for the LightGBM model for ``H1N1'' dataset. SHAP, SamplingSHAP, and ShapG rank the first four features in different orders, but the set of these features is the same: ``doctor recc h1n1 (10)'', ``opinion h1n1 risk (17)'', ``health insurance (15)'', and ``opinion h1n1 vacc effective (16)''. However, starting from the fifth feature, the feature rankings significantly differ for different XAI methods. Meanwhile, explanation of the FI method is significantly different from the results of other XAI methods. For the MLP model, we can see that SHAP, SamplingSHAP, LIME, and Permutation Feature Importance methods rank feature ``opinion h1n1 risk (17)'' as the most important feature, while this feature is ranked as the second by ShapG. Due to differences in explanations, it becomes extremely important to evaluate and compare these XAI methods. The comparison results are shown in Figures \ref{fig12} and \ref{fig14}.

\begin{remark}
 We should highlight that for LightGBM model we use Feature importance (FI) XAI method contrary to Perturbation Feature Importance (PFI) used for MLP model. The reason is as follows. FI is an explanation method built in tree models and it is widely used to explain tree-based models. However, for MLP models, which do not have tree structures to directly compute feature importance. Therefore, the impact of features in MPL models can be estimated using PFI  more efficiently. Since the PFI method is independent of the specific model, it can be widely used in various types of models to replace FI method in calculating feature importance. Thus, in our experiments we use FI in the LightGBM model and PFI in the MLP model for both datasets.
\end{remark}

\begin{table}[h!]
  \centering
  \caption{Feature importance ranking for ``H1N1'' dataset}\label{tab4}%
    \vspace{0.2cm}
  \textit{LightGBM model}
  \vspace{0.2cm}
  \begin{tabular}{|c|c|c|c|c|c|}
    \hline
    \diagbox{Rank}{XAI} & SHAP & SamplingSHAP & LIME  &Feature Importance& \textbf{ShapG}\\
    \hline
    Top 1& 15 &10 & 10  &30& 10\\   
    \hline
    Top 2& 10 & 15 & 15 & 35 & 17\\   
    \hline
    Top 3& 16 & 17 & 16 &34 &15\\   
    \hline
    Top 4& 17  & 16 & 14 &22& 16\\   
    \hline
    Top 5& 20 & 20 & 27 & 16 & 35 \\  
    \hline
    Top 6& 19 &22 & 20 & 17 &34\\  
    \hline
    Top 7& 11  & 30 & 17 & 31 & 14\\  
    \hline
    Top 8& 30 & 18 & 35 & 23 & 11 \\  
    \hline
    Top 9& 35  &35 & 18  & 20 &  20 \\  
    \hline
    Top 10& 21& 34  &4 & 26 & 2\\   \hline 
    \multicolumn{6}{c}{\,} \\
 \multicolumn{6}{c}{\textit{MLP model}}\\
 \hline 
    \diagbox{Rank}{XAI} &SHAP &SamplingSHAP & LIME  & Permutation Feature Importance & \textbf{ShapG}\\
    \hline
    Top 1 & 17  & 17 & 17  & 17 & 10\\   
    \hline
    Top 2& 10 &35 & 10 & 35 & 17\\   
    \hline
    Top 3 & 20 & 21 & 28 &10 &11\\   
    \hline
    Top 4  & 35 & 15  & 29 & 15 & 15\\   
    \hline
    Top 5 & 34 & 10 & 32 & 11 & 14 \\  
    \hline
    Top 6 & 15 & 30 & 22 & 31 & 16\\  
    \hline
    Top 7  & 16 &31 & 11 & 18 & 20\\  
    \hline
    Top 8 & 30 & 18 & 8 & 30 & 19 \\  
    \hline
    Top 9  & 27 & 22 & 30 & 29 & 35 \\  
    \hline
    Top 10& 1 & 29   &7 & 24 & 2\\   
    \hline
  \end{tabular}
\end{table}

Figures \ref{fig11} and \ref{fig13} show the changes in $R^2$ after gradually dropping features in different XAI methods  for LightGBM and MLP models, respectively, constructed for ``housing price'' dataset. From these data, we can intuitively compare the decrease and conclude that the accuracy in explanation of the results given by proposed XAI method ShapG is better than given by other existing methods.

Figures \ref{fig12} and \ref{fig14} show the changes in accuracy after gradually dropping features based on different XAI methods for LightGBM and MLP models, respectively, constructed for ``H1N1'' dataset. From Figure \ref{fig12} we can observe that in explanation results for the LightGBM model, the accuracy of three methods: SHAP, SamplingSHAP, and ShapG, exhibit a similar trend of steady decline. Although the accuracy of ShapG remains unchanged when removing the five and six features and other situations does not show an increase, it is noteworthy that SHAP shows an increase in accuracy when removing the first six and eight features, while SamplingSHAP shows continuous increase in accuracy when removing the first six, seven, and eight features. Moreover, although ShapG's explanation performance is comparable to most XAI methods, in terms of accuracy changes after the top ten features are removed, ShapG performs the best. As we can see in Figure \ref{fig14}, ShapG significantly outperforms other XAI methods in the explanation results based on the MLP model.

In order to compare running time of XAI methods based on game-theoretical approach, we compare KernelSHAP, SamplingSHAP, and ShapG. We perform the following results. 
Tables \ref{tab7} and \ref{tab8} show running time of  XAI methods for both LightGBM and MLP models on ``housing price'' and ``H1N1'' datasets, respectively. The comparison obviously shows that ShapG method is much faster among such game-theoretical methods as KernelSHAP and SamplingSHAP. With ``housing price'' dataset with 13 features, ShapG works more than 5 (4.8) times faster than KernelSHAP with LightGBM (MLP) model. We do not start KernelShap with ``H1N1'' dataset containing 35 features because an estimated time of its work is approximately 4219 (39284) min for LightGBM (MLP) model, which is more than 161 (685) times more than with our ShapG method.

\begin{table}[h!]
	\centering
	\caption{Running time (in sec) of XAI methods for ``housing price'' dataset}\label{tab7}%
	\vspace{0.2cm}
	\textit{LightGBM model}
	\vspace{0.2cm}
	
	\begin{tabular}{|c|c|c|c|}
		\hline
		XAI method & KernelSHAP& SamplingSHAP & \textbf{ShapG}\\
		\hline
		Running time &214.55 s  &42.14 s  &37.73 s\\
		\hline
		\multicolumn{4}{c}{\,} \\
		\multicolumn{4}{c}{\textit{MLP model}}\\
		\hline   
		XAI method & KernelSHAP& SamplingSHAP & \textbf{ShapG}\\
		\hline
		Running time &901.51 s &269.85 s &184.22 s \\
		\hline  
	\end{tabular}
\end{table}

\begin{table}[h!]
	\centering
	\caption{Running time of XAI methods for ``H1N1'' dataset}\label{tab8}%
	\vspace{0.2cm}
	\textit{LightGBM model}
	\vspace{0.2cm}
	
	\begin{tabular}{|c|c|c|c|}
		\hline
		XAI method & KernelSHAP& SamplingSHAP & \textbf{ShapG}\\
		\hline
		Running time & $\sim$ 4219 min & 374 min 87 s & 26 min 12 s \\
		\hline
		\multicolumn{4}{c}{\,} \\
		\multicolumn{4}{c}{\textit{MLP model}}\\
		\hline   
		XAI method & KernelSHAP& SamplingSHAP & \textbf{ShapG}\\
		\hline
		Running time & $\sim$ 39284 min &364 min 50 s & 57 min 22 s \\
		\hline 
	\end{tabular}
\end{table}

We do not consider SHAP in comparison analysis of a running time for several reasons:
\begin{itemize}
    \item When explaining tree models (e.g., LightGBM), SHAP calls the TreeSHAP Explainer. TreeSHAP utilizes the properties of the tree model itself to quickly compute approximate SHAP values, which significantly improves computational efficiency. However, the limitation of TreeSHAP is that it can only be used to explain tree models. This means that we cannot apply it to complex models such as neural networks.
    \item SHAP can automatically select the most suitable explainer for different AI models. When we use SHAP to explain MLP model, SHAP calls PermutationExplainer to generate explanation results. It works by iterating through forward and reverse feature permutations, which change the features one by one, thus effectively evaluating the independent contribution of each feature to the final output. PermutationExplainer does not directly construct a subset of all possible features, and does not use the Shapley value formula. Instead, it uses a method of permuting features to approximate the SHAP value.
\end{itemize}

By comparing evaluation results and running time, our XAI method ShapG provides a significant advantage in efficiency and speed. This further confirms the feasibility and usefulness of ShapG algorithm. Moreover, in the following section we show that it can be used for prediction models with complex architectures when other XAI methods based on game-theoretical approach, like KernelSHAP and SamplingSHAP do not work.\footnote{They could give the result but running time is so large that it makes them impossible to apply in practice, especially, for datasets with many features.}  This makes ShapG method a high-performance tool for XAI.

\subsection{Explanation of complex models}
\label{FI complex}

The proposed XAI method ShapG can be used not only for a single model like LightGBM or MLP as we show in the previous section, but it also provides global explanations for more complex models. These complex models include single models with complex architectures, ensemble learning models, hybrid models, etc. The models we use in this section are described in Section \ref{sec:models}. ShapG method can be applied to a wide range of models and can provide explanations of their decision-making processes. 

Tables \ref{tab11} and \ref{tab12} represent feature importance ranking given by ShapG for hybrid models constructed by combining different types of AI models in two-by-two combinations for ``housing price'' and ``H1N1'' datasets, respectively. We can observe very minor differences in the ranking of feature importance for different models.

\begin{table}[h!]
	\centering
	\caption{Feature importance ranking by ShapG for complex AI models for ``housing price'' dataset}\label{tab11}
	\begin{adjustbox}{width=\textwidth}
		\begin{tabular}{|c|c|c|c|c|c|c|c|c|}
			\hline
			\diagbox{Rank}{AI Model} & LGB  & Stacking  & Linear-LGB & Linear-MLP & Linear-KNN & MLP-LGB &KNN-LGB  & MLP-KNN\\
			\hline
			Top 1 & 13  & 13  & 13 & 13 & 13 & 13 &13  & 13 \\
			\hline
			Top 2& 6 & 5  & 6& 6 & 6 & 6 & 6  & 6\\
			\hline
			Top 3& 5 & 6  & 5 & 2 & 11 & 3 &5  & 3\\
			\hline
			Top 4& 11 & 3  & 11 & 3 & 3 & 5 &3  & 11\\
			\hline
			Top 5& 3  & 11  & 3 & 11 & 5 &2 &11  & 2 \\
			\hline
			Top 6& 2  & 2  & 2 & 8 & 2 & 11 &2 & 5\\
			\hline
			Top 7& 8  & 10  & 8 & 1 & 10 & 8 &8  & 1\\
			\hline
		\end{tabular}
	\end{adjustbox}
\end{table}

For the ``housing price'' dataset, from Table \ref{tab11} we can clearly observe that feature ``LSTAT -- lower status of the population (13)'' is the most important feature in predicting housing prices, and it is the same as is identified by simple LightGBM and MLP models. In addition, we can also notice that features ``RM -- average number of rooms per dwelling(6)'', ``INDUS -- proportion of non-retail business acres per town (3)'', and ``NOX -- nitric oxides concentration (5)'' appear more frequently in the top important features among all complex models. 

For the ``H1N1'' dataset, from Table \ref{tab12} we can clearly observe that ``doctor recc h1n1 (10)'' and ``opinion h1n1 risk (17)'' are the two most important features, which indicate that they have a significant effect on whether respondent received H1N1 flu vaccine. Similarly, it can be noticed that ``health insurance (15)'' and ``opinion h1n1 vacc effective (16)'' appear more frequently among the most important features. This means that these features are also important factors that affect people's intention to receive vaccination.  

\begin{table}[h!]
  \centering
  \caption{Feature importance ranking by ShapG for complex AI models for ``H1N1'' dataset}\label{tab12}
    \begin{adjustbox}{width=\textwidth}
     \begin{tabular}{|c|c|c|c|c|c|c|c|c|}
      \hline
       \diagbox{Rank}{AI Model} & LGB  & Stacking  & Logistic-LGB & Logistic-MLP & Logistic-KNN & MLP-LGB &KNN-LGB  & MLP-KNN\\
      \hline
      Top 1 & 10  & 10  & 10 & 10 & 10 & 10 &10  & 10 \\
     \hline
      Top 2& 17 & 17  & 17 & 17 & 17 & 17 & 17  & 17\\
      \hline
     Top 3& 15  & 15  & 16 & 15 & 20 & 15 &16 & 15\\
      \hline
      Top 4& 16 & 16  & 11 & 14 & 16 & 11 &15  & 14\\
      \hline
      Top 5& 35  & 35  & 15 & 11 & 15 &14 &35 & 11 \\
      \hline
      Top 6& 34  & 14  & 14 & 20 & 11 & 35&11 & 24\\
      \hline
      Top 7& 14  & 34  & 20 & 16 & 14 & 16 &34 & 13\\
      \hline
      Top 8& 11  & 11  & 19 & 5 & 24 & 20 &14 &20\\
      \hline
      Top 9& 20 & 20  & 13 & 13 & 13 & 34 &20  & 16 \\
      \hline
      Top 10& 2 & 19  & 1 & 1 & 29 &24 &2  &35 \\
      \hline
    \end{tabular}
  \end{adjustbox}
\end{table}

\begin{table}[h!]
	\centering
	\caption{Running time by ShapG for complex AI models for ``housing price'' dataset}\label{tabRTHousing}
	\begin{adjustbox}{width=\textwidth}
		\begin{tabular}{|c|c|c|c|c|c|c|c|c|}
			\hline
			AI Model& LGB  & Stacking  & Linear-LGB & Linear-MLP & Linear-KNN & MLP-LGB &KNN-LGB  & MLP-KNN\\
			\hline
			Running time & 37.73 s  & 457.37 s  & 51.51 s & 364.32 s & 11.03 s & 536.28 s &51.47 s  & 377.75 s \\
			\hline
		\end{tabular}
	\end{adjustbox}
\end{table}

\begin{table}[h!]
  \centering
  \caption{Running time by ShapG for complex AI models for ``H1N1'' dataset}\label{tabRTH1N1}
    \begin{adjustbox}{width=\textwidth}
     \begin{tabular}{|c|c|c|c|c|c|c|c|c|}
      \hline
       AI Model & LGB  & Stacking  & Logistic-LGB & Logistic-MLP & Logistic-KNN & MLP-LGB &KNN-LGB  & MLP-KNN\\
      \hline
      Running Time & 26 min 12 s & 249 min 35s  & 42 min 31 s & 61 min 17 s & 115 min 02 s & 105 min 03 s &157 min 16 s  & 156 min 44 s \\
      \hline
    \end{tabular}
  \end{adjustbox}
\end{table}

Tables \ref{tabRTHousing} and \ref{tabRTH1N1} show the running time required by the ShapG algorithm to explain complex AI models for ``housing price'' and ``H1N1'' datasets, respectively. Depending on the size of the dataset and model complexity, ShapG requires different running time. As expected, running time for ``H1N1'' dataset containing 35 features is much larger than the time required for the same model constructed for ``housing price'' dataset with 13 features.  Although running time required by ShapG for complex AI models is much larger than for simple models (LightGBM or MPL), the ShapG method has many significant advantages in comparison with other XAI methods. The ShapG method has not only been applicable to any model from theoretical point of view, but its performance is also verified in many experiments with complex AI models. Although SHAP, SamplingSHAP, and KernelSHAP are assumed to be applicable to explain any model, during our experiments we found that when applying these methods to more complex neural network models or hybrid models, the code often did not run successfully and did not give valid explanation results. These methods lack good compatibility with complex AI models and require more in-depth adjustments to the dataset or AI models. Therefore, for researchers who are not specialized in the field of AI, these methods cannot easily provide explainable results. When explaining complex AI models, ShapG is still able to provide reliable explanation results. 

\section{Conclusions}\label{sec6}

In this paper, we proposed a new explainable artificial intelligence (XAI) method called ShapG, which is based on Shapley value for graph games. It is a model-agnostic global explanation method. ShapG calculates feature importance by constructing an undirected graph of features, where nodes in the graph represent features, and samples based on graph. It starts with an empty graph and consequently adds the edges, which are pairs of features with the strongest correlation. The algorithm stops when all features are connected to ensure that the feature graph contains important structural information. In the process of calculating the Shapley value, we only need to consider the coalitions between each node and its neighbors, not all possible coalitions. This optimization improves the efficiency of the algorithm.

We have compared ShapG with several popular XAI methods, e.g., Feature Importance (FI), Permutation Feature Importance (PFI), LIME, SHAP, SamplingSHAP, and KernelSHAP. Our ShapG exhibits excellent explanation results, which are significantly better than other XAI methods for two datasets. In addition, compared to SamplingSHAP and KernelSHAP methods also based on cooperative game theory, ShapG saves significant computational resources in running time. These results provide validation of reliability and wide applicability of our method.

We believe that ShapG can be considered as a useful XAI method that can be applied not only to simple AI models, but also to provide global explanations for complex models. It can reliably explain decision-making process of complex models, thus helping users to better understand these models. We will continue to study how to optimize our algorithm and reduce its running time to provide users quicker and more trustworthy explanation results.

\bibliographystyle{unsrtnat}
\bibliography{references}  






\end{document}